\documentclass[journal]{IEEEtran}

\usepackage{gensymb}

\usepackage{multirow}

\usepackage{subfigure}

\usepackage{bm}

\usepackage{cite}

\usepackage{color}

\ifCLASSINFOpdf
   \usepackage[pdftex]{graphicx}
   \graphicspath{{../pdf/}{../jpeg/}}
   \DeclareGraphicsExtensions{.pdf,.jpeg,.png}
\else
   \usepackage[dvips]{graphicx}
   \graphicspath{{../eps/}}
   \DeclareGraphicsExtensions{.eps}
\fi

\usepackage{amsmath}

\usepackage{color}

\begin{document}

\title{Accurate Object Association and Pose Updating \\for Semantic SLAM}

\author{Kaiqi~Chen,~Jialing~Liu,~Qinying~Chen,~Zhenhua~Wang,~Jianhua~Zhang

\thanks{K. Chen, J. Liu, Q. Chen and Z. Wang are with the Institute of Computer Vision, College of Computer Science and Technology, Zhejiang University of Technology, Hangzhou 310023, China (e-mail: chenkaiqi96@outlook.com; liujialing98@hotmail.com; chenqy@zjut.edu.cn; zhhwang@zjut.edu.cn).}
\thanks{J. Zhang is the corresponding author. He is with the School of Computer Science and Engineering, Tianjin University of Technology, Tianjin 300384, China (e-mail: zjh@ieee.org).}

}

\markboth{Journal of \LaTeX\ Class Files,~Vol.~14, No.~8, August~2015}%
{Shell \MakeLowercase{\textit{et al.}}: Bare Demo of IEEEtran.cls for IEEE Journals}

\maketitle

\begin{abstract}
Current pandemic has caused the medical system to operate under high load.
To relieve it, robots with high autonomy can be used to effectively execute contactless operations in hospitals and reduce cross-infection between medical staff and patients. Although semantic Simultaneous Localization and Mapping (SLAM) technology can improve the autonomy of robots, semantic object association is still a problem that is worthy of being studied.
The key to solving this problem is to correctly associate multiple object measurements of one object landmark by using semantic information, and to refine the pose of object landmark in real time.
To this end, we propose a hierarchical object association strategy and a pose-refinement approach. The former one consists of two levels, \textit{i.e.}, a short-term object association and a global one. In the first level, we employ the multiple-object-tracking for short-term object association, through which the incorrect association among objects whose locations are close and appearances are similar can be avoided. Moreover, the short-term object association can provide more abundant object appearance and more robust estimation of object pose for the global object association in the second level. To refine the object pose in the map, we develop an approach to choose the optimal object pose from all object measurements associated with an object landmark.
The proposed method is comprehensively evaluated on seven simulated hospital sequences\footnote{Link: https://pan.baidu.com/share/init?surl=dPgPD1RUi6F9zeWSXicVNw (extraction code: 8ocw)}, a real hospital environment and the KITTI dataset. 
Experimental results show that our method has an obviously improvement in terms of robustness and accuracy for the object association and the trajectory estimation in the semantic SLAM.\\\end{abstract}


\IEEEpeerreviewmaketitle

\section{Introduction}

\IEEEPARstart{U}{nder} the current serious pandemic, a large number of autonomous robots are urgently needed to assist medical staff in their work \cite{yang2020combating}, such as delivering supplies, so as to relieve the pressure of the medical system. Although the precise service of intelligent medical robots has been getting more and more attention in the past few decades, medical robots currently do not have enough intelligence and autonomy to assist doctors and nurses. Realizing intelligent guidance and emergency assistance through the vision-assisted medical robot is a crucial but difficult task in the intelligent medical field. The vision-assisted medical robot can use visual navigation and localization technology to accurately locate landmarks and quickly deliver medical supplies. In such a robot system, the visual Simultaneous Localization and Mapping (SLAM) plays an important role to improve its autonomies. 

Traditional visual SLAM uses very little semantic information in localization and mapping, so that it is restricted in some application scenarios. Although the visual SLAM has general robust performance, it is easy to lose in dynamic scenes, fast motion, texture loss, lighting changes and other situations. With the reform of the medical service system, traditional SLAM cannot meet the increasing demand. Some researchers have proposed several semantic SLAM systems, \textit{e.g.}\cite{bowman2017probabilistic, mu2016slam, salas2013slam++, rosinol2020kimera, choi2016local}, by combining traditional SLAM with semantic information to improve the robustness of the system, which is also more consistent with human cognition of exploring unknown environments.
In the semantic SLAM system, the key is to make use of objects as landmarks. Thus, the correct object association can improve the robustness and precision of SLAM, and SLAM can also provide a better initial value for object association. These two aspects are tightly coupled. However, semantic information has not been fully utilized in their work. Moreover, their object associations are performed in the entire environment, so all object measurements are considered at the same time. 
As the running time and map scale increase, it is inevitable to detect more object measurements. Thus, associating all these object measurements at the same time requires more computation resources and may lead to a decrease in the accuracy of the association, which is obviously inefficient.

\begin{figure}[t!]
\includegraphics[width=3.5in, height=1.75in] {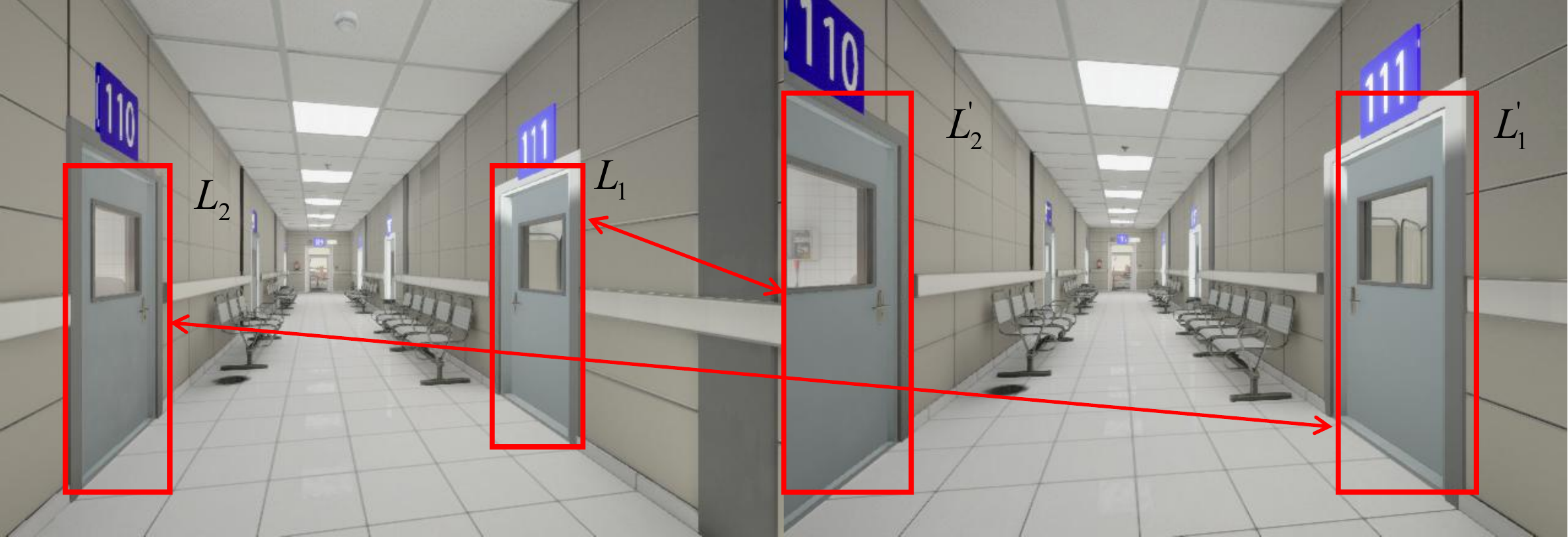}
\caption{The examples of simulated hospital taken by UAVs. The two pictures are taken at different times. The left picture is detected with two object measurements $L_{1}$, $L_{2}$ at time $t$, while the right picture is detected with two object measurements $L_{1}^{'}$, $L_{2}^{'}$ at time $t+1$. Assuming that $L_{1}$, $L_{2}$ is observed for the first time at time $t$, then $L_{1}^{'}$, $L_{2}^{'}$ will be associated with $L_{1}$, $L_{2}$ at time $t+1$. Since $L_{1}$ and $L_{2}$ are very close and very similar, it is easy that $L_{1}^{'}$ is incorrectly associated with $L_{2}$, and $L_{2}^{'}$ is wrongly associated with $L_{1}$. } 
\label{Fig.1.} 
\end{figure}

In order to improve the accuracy of object association and ensure the robustness of SLAM, a Hierarchical Dirichlet Processing (HDP)-based object association method is proposed and the corresponding semantic SLAM system is implemented in our previous work\cite{zhang2019hierarchical}. This method treats each keyframe as a group, object measurements in the same group cannot be associated with the same object landmark. For each object measurement, only the keyframes that have a common view with the keyframe possessing this object measurement have to be traversed to determine whether to associate the same object landmark or associate with a new object landmark. Through such an association, the method achieves better accuracy for object association and higher robustness for the system. However, this method solely utilizes the position and appearance of each single object measurement in world coordinates to calculate the probability that determines whether an object measurement is the same as other object measurements.
Thus, when multiple objects in the same class are close together, it is very challenging to correctly associate object measurements, as shown in Fig.\ref{Fig.1.}.

Furthermore, how to select the best pose estimation for an object landmark is also a problem that has not been carefully considered by \cite{zhang2019hierarchical}, where the pose of the object landmark comes from the pose of the first object measurement associated with this landmark.
Because an object landmark can be associated with many object measurements, and the estimated pose of the first one may not be accurate, it is obviously unreasonable to use its pose as the pose of object landmark.

To achieve robust object association and accurate estimation of object pose, we propose a hierarchical object association strategy based on keyframe groups, and an approach to choose an optimal pose for an object landmark from all object measurements associated with it. The hierarchical object association strategy consists of two levels. In the first level, the object association is executed within a keyframe group formed by several adjacent keyframes, where a Multiple Object Tracking (MOT) method\cite{zhang2021fairmot} is employed for short-term object association to avoid incorrect association among close and similar objects. In the second level, the HDP model is used for global object association. With the help of short-term object association, we can obtain more abundant appearance features and more robust pose estimations, which further improve the correctness of global object association.

\begin{figure}[ht]
\includegraphics[width=3.5in, height=2in] {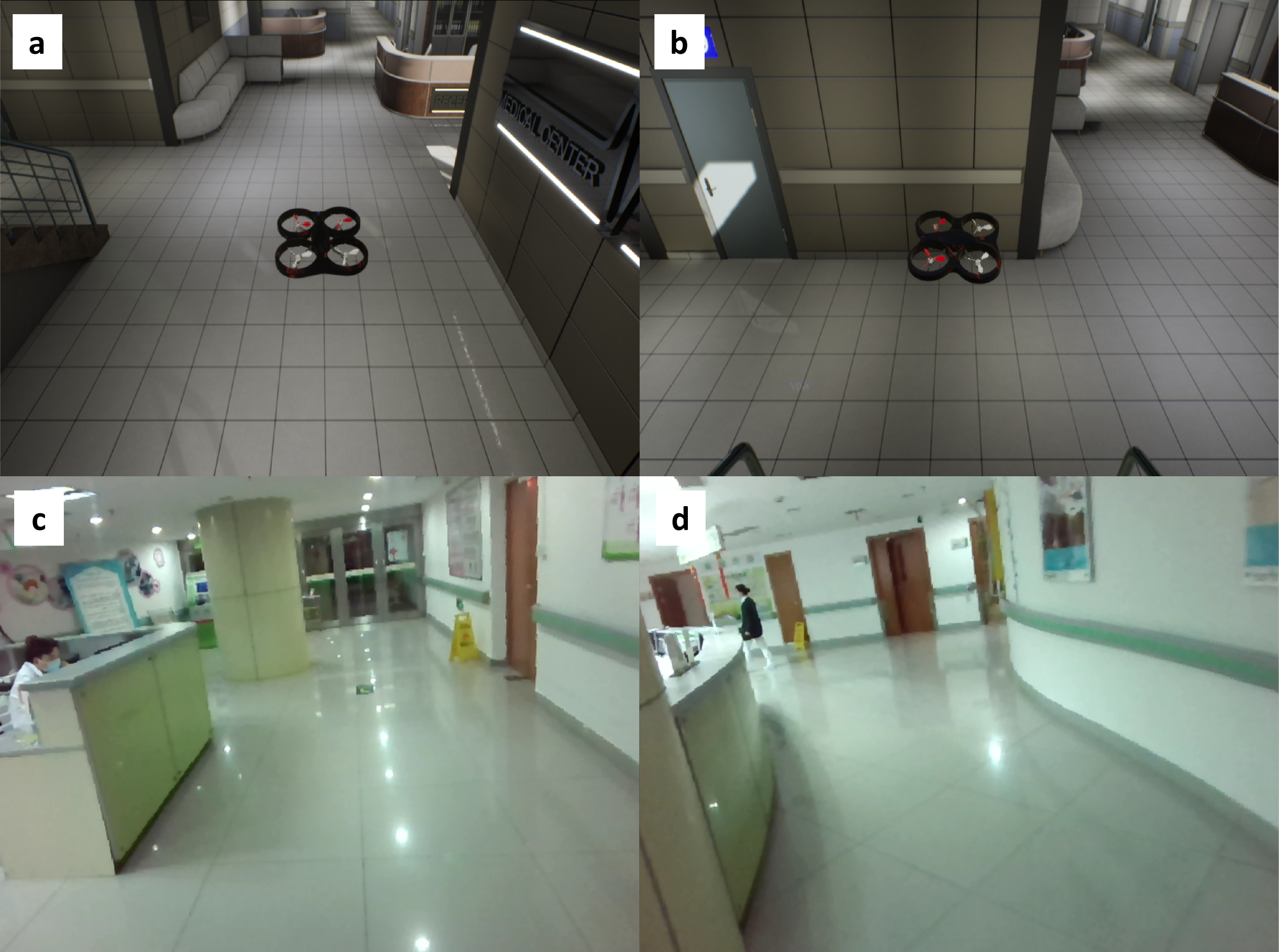}
\caption{Examples of four scenes are taken from the simulated hospital (the upper row) and real hospital (the bottom row).}
\label{Fig.2.} 
\end{figure}

We carry out extensive experiments to verify the performance of the proposed method in a simulated hospital constructed by AirSim\cite{shah2018airsim}, from which we collect several sequences from viewpoints of Unmanned Aerial Vehicle (UAV), as shown in Fig.\ref{Fig.2.}. We also evaluate the proposed method in three real hospital sequences, as shown in Fig.\ref{Fig.2.}, and some real sequences in the KITTI datasets\cite{geiger2013vision}. Experimental results show that our method is more accurate and robust than state-of-the-art (SOTA) semantic SLAM systems in terms of object association and trajectory errors.

In summary, our main contributions are summarized as follows:

\begin{itemize}

\item[$\bullet$] A hierarchical object association algorithm is proposed based on a two-level strategy, through which the association errors caused by objects with close positions and similar appearances can be avoided, and consequently a more accurate trajectory and semantic map can be obtained by the semantic SLAM.

\item[$\bullet$] A refinement approach to choose optimal object poses is proposed to improve the estimation of object pose, and consequently enhance the accuracy of semantic maps.

\item[$\bullet$] Experiments carried out on simulated and real hospitals, as well as sequences in the KITTI dataset, demonstrate the superiority of our approach over existing SOTA methods in terms of association accuracy and trajectory errors.

\end{itemize}

\section{Related Work}

After decades of development of SLAM, SLAM has achieved amazing progress, especially in recent years. As the camera can be easier installed on different equipments and can obtain richer environmental information, there are many mature visual SLAM systems widely used in various scenarios. From the perspective of visual SLAM methods, it can be divided into direct methods \cite{engel2014lsd, forster2014svo} and feature-based methods \cite{klein2007parallel,mur2015orb}. The typical direct methods, \textit{e.g.} the LSD-SLAM\cite{engel2014lsd}, are based on direct image alignment and use accurate motion estimation to reconstruct the environment. But direct methods are easily influenced by dramatical light change.

As one of the typical feature-based SLAM, the ORB-SLAM\cite{mur2015orb} is more robust to light changes, but it easily fails to track under low-texture conditions. Subsequently, the ORB-SLAM2\cite{mur2017orb} is proposed to improve accuracy and robustness of the original ORB-SLAM, by adding the calibrated Stereo camera and RGB-D camera. Recently, Campos \textit{et al.}\cite{campos2021orb} further improve it and propose ORB-SLAM3, which is very suitable for long-term and large-scale SLAM practical applications. 

Many SLAM systems have been successfully used by intelligent medical service robot systems. For instance, some researchers have achieved good results in medical surgery\cite{qiu2018endoscope, marmol2019dense} and intelligent medical service robots\cite{wang2007multi, civera2011towards, fang2021visual}. The methods proposed by Qiu \textit{et al.}\cite{qiu2018endoscope} and Marmol \textit{et al.}\cite{marmol2019dense} are based on the ORB-SLAM\cite{mur2015orb}. Both of them perform precise localization and 3D reconstruction for a certain part of the body, which provide detailed and accurate information for the surgeon to observe the painful part of the patient. Furthermore, some researchers are extending SLAM to autonomous driving. Chen \textit{et al.}\cite{chen2020event} points out a new research direction for semantic SLAM by using bio-inspired sensors such as event cameras, which are characterized by low energy consumption and low latency.

In the early research stage of intelligent medical service robots, Wang \textit{et al.}\cite{wang2007multi} provide a method to fuse the local maps into a jointly maintained global map by first transforming the local map state estimation into relative location information, and then conducting the fusion using the decoupled D-SLAM\cite{wang2007d} framework. So that multiple robots can find the map overlap and corresponding landmarks on the map.

Although visual SLAM has achieved great progress, traditional visual SLAM only uses meaningless feature points as landmarks, which are not robust under poor environmental conditions.
In recent years, the combination of semantic information and SLAM has become more and more prevalent. Semantic information combined with SLAM is mainly used in two modules, localization\cite{mur2017orb, campos2021orb} and mapping\cite{lianos2018vso, mccormac2018fusion++}. By using the semantic information to assist the system localization, a relatively accurate object pose is obtained. Civera \textit{et al.}\cite{civera2011towards} propose a semantic SLAM algorithm that merges traditional meaningless points with known objects in the estimated map. When the camera is installed on a mobile robot and moves in the hospital scene, the robot performs a grasping task and the algorithm can also run in real time.

The key to using semantic information is object association. Common object association can be mainly achieved by the following methods, Extended Kalman Filter (EKF)\cite{neira2001data}, Expectation Maximization (EM)\cite{bowman2017probabilistic, strecke2019fusion}, Non-parametric Probability Model (NPM)\cite{mu2016slam, zhang2019hierarchical} and Deep Learning (DL) methods\cite{xiang2017rnn, liu2019monocular, yang2019cubeslam}, etc. EKF uses the odometer to obtain the landmark information at the current moment, matches with the previous landmarks, obtains the information of the robot, re-estimates its own position, and finally updates the location and uncertainty of each landmark. After SLAM initialization, the incorrect pose estimation or the fuzzy mapping may easily lead to problems in object association.

The EM algorithm is employed by Bowman \textit{et al.} in \cite{bowman2017probabilistic} and Strecke and Stuckler in \cite{strecke2019fusion} to deal with object association problems. Bowman \textit{et al.}\cite{bowman2017probabilistic} tightly couple inertial, geometric and semantic observations into a single optimization framework, verifying the performance of the algorithm on indoor and outdoor datasets. Later, the EM-Fusion \cite{strecke2019fusion} also uses the EM algorithm for an object-level dense dynamic SLAM, which has greatly improved the accuracy of object association and can handle object occlusion.

Xiang \textit{et al.}\cite{xiang2017rnn} propose to use the recurrent neural network to solve the object association problem. They provide semantic tags for objects and use recursive units to connect information on multiple views. Liu \textit{et al.}\cite{liu2019monocular} use the most advanced convolutional neural networks (CNNs) and Structure From Motion (SFM) for object detection and scene reconstruction, which can be used to identify repetitively computed scenes, and improve object association. 
Yang \textit{et al.}\cite{yang2019cubeslam} adopt an efficient, accurate and robust single-image 3D cuboid detection method. In terms of object association, they use two methods to associate objects. For static objects, they associate object measurements based on feature point matching. For dynamic objects, they utilize the visual object tracking algorithm.
Bavle \textit{et al.}\cite{bavle2020vps} present a lightweight and real-time visual semantic SLAM framework running on UAV platforms. They use the constraints between detected semantic planar surfaces to associate objects.
Qin \textit{et al.}\cite{qin2020avp} build a map for self-driving vehicles in parking lots by means of the predicted parking lines for semantic object association. However, there are no semantic objects in the map and the pose of parking lines are used for optimization.
Li \textit{et al.}\cite{li2020textslam} use the detected text as the semantic object to build a semantic mapping. Thus, the object association can be easily achieved by the text recognition. However, this association can only be used in very limited environments.
Qian \textit{et al.}\cite{qian2021semantic} use a novel object-level object association algorithm based on the bag of words algorithm. In dealing with the object association problem, they used the geometry and appearance information of the object.
Chen \textit{et al.}\cite{chen2021pole} provide a novel pole-curb fusion localization system. They select poles and curbs as landmarks, and adopt a Branchand- Bound (BnB)-based global optimization method to tackle the data association problem of poles.
Although these semantic SLAM systems have achieved impressive results, the object association in these systems is far from mature, as there are still many association errors when the real environment is large and complex.

To enhance the object association, the Dirichlet Process is introduced by Mu \textit{et al.}\cite{mu2016slam} to enjoy the advantage without determining the number of object landmarks beforehand. 
Inspired by this work, we further employ the Hierarchical Dirichlet Process to achieve the hierarchical object association in our previous work \cite{zhang2019hierarchical}. It considers each keyframe as a document, each object landmark as a topic, and object measurements as words forming the document. Thus, the object association can be modeled as an HDP model and the grouped data (\textit{i.e.}, object measurements) can be clustered in a hierarchical style. Different from \cite{mu2016slam}, the HDP-SLAM associates object measurements in one keyframe with those object landmarks that are observed in this keyframe, while in \cite{mu2016slam}, the method associates object measurements with all object landmarks. Moreover, the HDP-SLAM tightly couples object association and pose optimization, where the object pose is also involved in the process of Bundle Adjustment (BA).
However, when multiple objects of the same category are close together in a complex environment, the HDP-SLAM may wrongly associate object measurements.

To tackle this problem, we propose to use the FairMOT \cite{zhang2021fairmot} to associate object measurements in several adjacent keyframes. As the MOT has always been a long-term goal in the field of computer vision, many methods have been proposed \cite{lee2016ground, sanchez2016online, bewley2016simple, wojke2017simple, sun2019deep, pang2020tubetk}.
Among these methods, the FairMOT \cite{zhang2021fairmot} can achieve superior performance by comparing with DeepSORT\_2\cite{wojke2017simple}, SST\cite{sun2019deep}, and TubeTK\cite{pang2020tubetk} in different test datasets. Although the FairMOT is the one-step MOT, which is not as good as the two-step method in terms of object detection and tracking accuracy, it still has the advantage in real-time. In the one-step method, the detection model and association model can share most of the features, reducing the processing time, which makes it more suitable for the SLAM systems, in which the requirement of real-time is very critical.

\begin{figure*}[htbp]
\centerline{\includegraphics[width=1.0\textwidth]{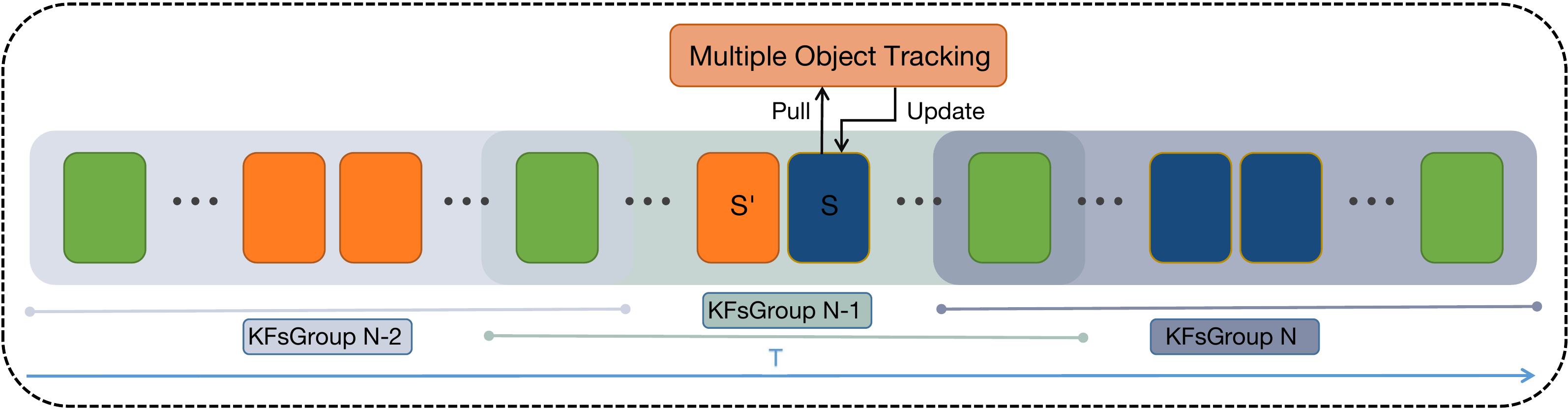}}
\caption{Illustration of keyframe groups. In timeline T, the keyframe $S$ (blue) is entering multi-object tracking. The keyframe $S^{'}$(orange) has been detected and tracked, and the object information has been acquired. The keyframe queue is divided into $N$ keyframe groups, and adjacent keyframe groups have $j$ overlapping keyframes (green).}
\label{Fig.3.}
\end{figure*}

\begin{figure}[ht]
\centerline{\includegraphics[width=0.4\textwidth]{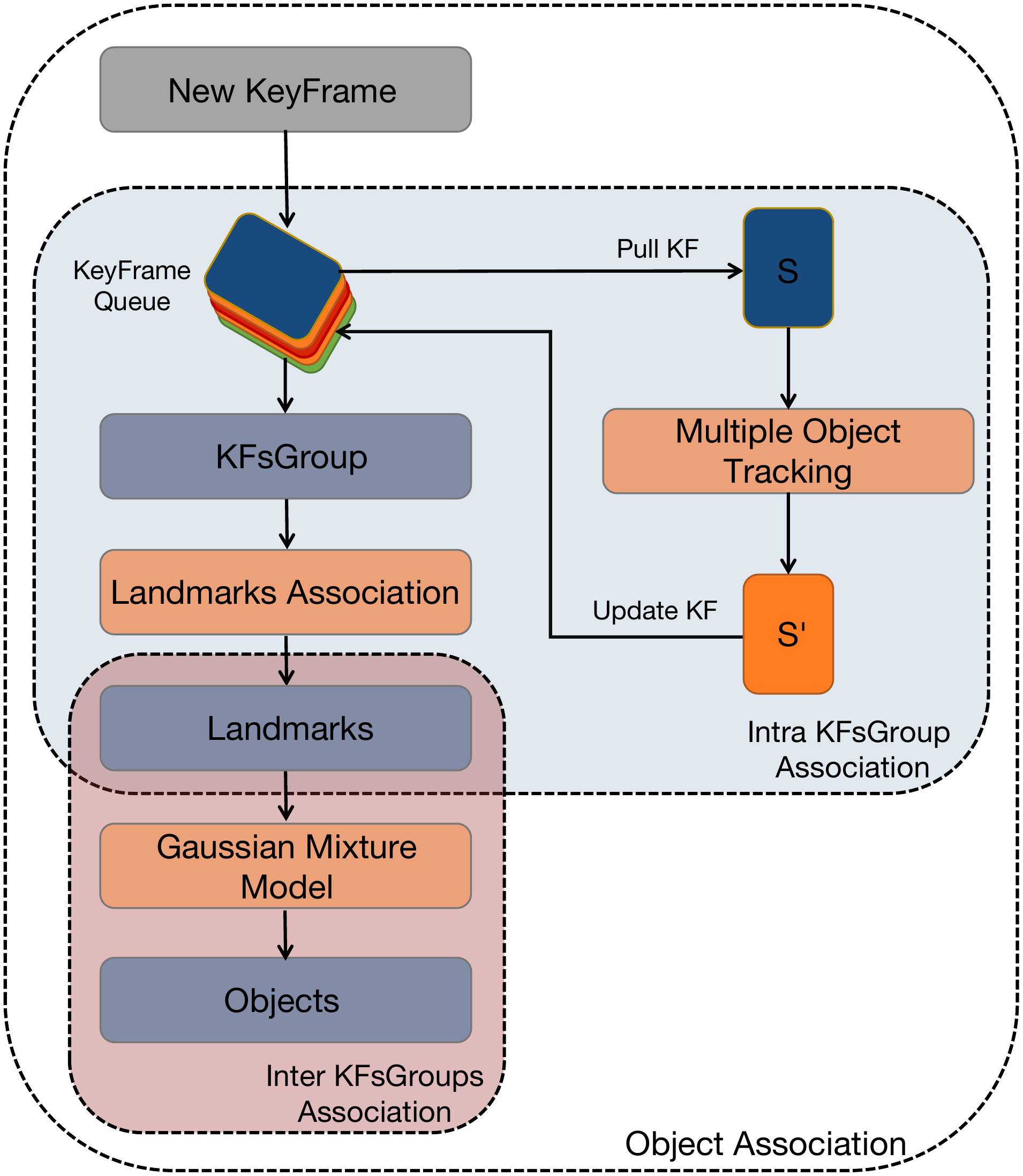}}
\caption{Hierarchical object association process. After each keyframe is added to the keyframe queue, intra and inter groups object association will be handled. Intra Group Association: the blue keyframe turns into an orange keyframe after multi-object detection and tracking, and the orange keyframe contains object measurements. Inter Group Association: the objects in the keyframe group are processed to obtain the final real objects with the help of GMM and HDP.}
\label{Fig.4.}
\end{figure}

\section{Object SLAM}
In the original HDP-SLAM method\cite{zhang2019hierarchical}, the object measurement is obtained through deep learning method, \textit{i.e.}, the SSD model\cite{liu2016ssd}. It detects objects and outputs 2D bounding boxes from each keyframe. Then, image patches cropped according to these bounding boxes are input into a pose prediction network\cite{tulsiani2015viewpoints} to obtain the 3D pose of object measurements. This processing does not take into account that the spatiotemporal correlation of object measurements in several adjacent keyframes. Therefore, this processing has two limitations. The first is that object measurements with similar appearances and close positions observed from several adjacent keyframes may be wrongly associated. The second is that object measurements may be lost as some objects cannot be detected.

Inspired by the observation that one object can be tracked in a short time with very high accuracy, in this study, we propose to use one-step MOT\cite{zhang2021fairmot} to detect and track object measurements in several adjacent keyframes. We group these adjacent keyframes. Thus, in one group of keyframes, object association can be achieved by object tracking. Because the number of keyframes in this group is small (it is usually less than 10), object tracking is precise and object association within the group is accurate. To associate object measurements among groups, the HDP is also employed as used in HDP-SLAM\cite{zhang2019hierarchical}. After object association among groups, the pose of each object landmark will be refined by considering all poses of object measurements associated with this landmark.

\subsection{Object Association}\label{3.1}
In the HDP-SLAM, each keyframe is treated as a group, and the object measurement in each new keyframe is directly associated with the object landmark. It ignores the strong spatiotemporal correlation of object measurements that belong to one object landmark but are detected from several adjacent keyframes. Thus the incorrect association may occur when similar and close objects are detected in adjacent keyframes. In order to improve the correctness of object association, we propose a hierarchical strategy consisting of two-level object association.

As shown in Fig.\ref{Fig.3.}, the SLAM system selects representative frames from all frames as keyframes $F_{1:D} =\left \{ F_{1} ,.. .,F_{D} \right \}$ to reduce redundant computation costs at time $t$ in the time axis $T$. For adjacent $M$ keyframes, we group them as a keyframe group. At the time $t$, we assume there are totally $N$ groups. Two adjacent keyframe groups have $j$ keyframes overlap, which adds a constraint to the object association between keyframe groups. Thus, we denote the $G_t = \left \{   g_{1},...,g_{N} \right \} $ as the keyframe groups we obtained at $t$ time.

In each keyframe group, we associate object measurements through the FairMOT algorithm\cite{zhang2021fairmot}. Because there are only $M$ adjacent keyframes, where $M$ is usually less than 10, the object measurements belonging to one object landmark can be detected and tracked with high accuracy. For object association among keyframe groups, we still utilize HDP as in \cite{zhang2019hierarchical}. In this study, we can further build a Gaussian Mixture Model (GMM)\cite{reynolds2009gaussian} to represent the object landmark associated with multiple object measurements within one keyframe group. Through this model, a more accurate prior can be computed when executing object association among keyframe groups through Gibbs sampling in the HDP. Finally, a more precise object association can be achieved.

\subsubsection{Object Association within a Single Keyframe Group}

In a keyframe $F_m$, a set of object measurements $L_m=\{L_{m,1}, \cdots, L_{m,K_m}\}$ can be obtained, where $K_m$ means the number of object measurements that are detected and tracked in $F_m$. An object measurement $L_{m,k}$ contains all information required by the following object association,

\begin{equation}
L_{m, k}= \left \{ ID_{m, k}^{obj},ID_{m, k}^{kf},Box_{m, k},Pose_{m, k} \right \}, \label{3}
\end{equation}
where $ID_{m, k}^{obj},ID_{m, k}^{kf},Box_{m, k}$, and $Pose_{m, k}$ are the detected object ID, the keyframe ID, the detected bounding box information, and the object pose, respectively.

In general, one object landmark can be observed as multiple object measurements in several adjacent keyframes. Thus, several object measurements of one object landmark can be detected in a keyframe group. Through the FairMOT, these object measurements can be efficiently and accurately associated  with one object landmark, as it can simultaneously track multiple object instances across several keyframes. In the $n$th keyframe group, we denote the association between the $k$th object measurement in the $m$th keyframe (\textit{i.e.} $L_{m,k}^G$) and the $i$th object landmark as $s_{n,i}^{m,k}=<O_{n,i}^G,L_{m,k}^G>$, where the superscript $G$ denotes the association is within one keyframe group. If an object measurement is detected and cannot be associated with any previous object landmark, a new object landmark is generated. Finally, we have a set of object landmarks in the level of intra-keyframe-group, denoted as $O_n^G=\{O_{n,1}^G, \cdots, O_{n,I}^G\}$. It should be noted that these object landmarks are not the final object landmarks used in the global map, but will be associated in the global level. This is because we design a hierarchical group strategy, and consequently the same hierarchical object association.

For each object landmark $O_{n,i}^G$ in the keyframe group, there are usually more than one object measurements associated with it. Thus, we can describe the object landmark in a more precise manner, through which the object association among keyframe groups will be more accurate and robust.

\subsubsection{Object Association among Consecutive KeyFrame Groups} 

The HDP-SLAM uses HDP to associate objects among groups where each group is just a keyframe. The priors required by HDP inference are computed from the difference of location and appearance between the object measurement and the object landmark. This approach is easily affected by inaccurate measurements and changes of environmental and lighting conditions, resulting in some wrong associations. In this study, the basic element of object association between keyframe groups is no longer a single object measurement detected in each keyframe, but the object landmark, $O_{n,i}^G$, associated by object measurements within each keyframe group. Because the object landmarks in a keyframe group are usually associated with multiple object measurements, it can provide more pose and appearance information for landmarks in such a group, and thus it is more robust to inaccurate measurement and condition changes.

In this study, we employ the GMM\cite{reynolds2009gaussian} to model the pose information for multiple object measurements associated with one object landmark in a keyframe group.

\begin{equation}
{\Gamma}   \left ( O\mid \Theta   \right ) =\sum_{z=1}^{Z} \lambda_{z}\phi \left ( O\mid \Theta_{z} \right ) ,
\label{equ.7}
\end{equation}
where $\lambda_{z}$ is the weight coefficient, $\phi \left (O\mid \Theta_{z} \right) $ is the Gaussian distribution density, $\Theta_{z}=\left \{ \bm{\mu}_{z},\bm{\Sigma}_{z} \right \}$, and 

\begin{equation}
\phi \left ( O\mid \Theta_{z} \right ) =\frac{1}{\sqrt{2\pi }|\bm{\Sigma}_{z}|^{\frac{1}{2}} } \bm{e}^{ -\frac{\left (O-\bm{\mu}_{z}\right )^{\text{T}}\bm{\Sigma}^{-1}_{z}(O-\bm{\mu}_{z})}{2} } .
\label{equ.8}
\end{equation}

In Eq. \eqref{equ.7}, the object landmark $O$ is associated with $Z$ object measurements, and the observed data of each measurements is a set with six free variables, namely the 3D world position and the orientation of the three axes of the object.

As described in HDP-SLAM\cite{zhang2019hierarchical}, a prior of an object measurement should be computed according to the proportion of distance between object measurements, through which the object landmark that current object measurement is belonging can be determined during Gibbs sampling of HDP.
In this method, we improve this prior probability by using the above GMM. When determining if the $i$th object landmark $O_{n,i}^G$ in the $n$th keyframe group can be associated with the $j$th object landmark $O_{m,j}^G$ in the $m$th keyframe group, we compute the probabilities of all object measurements associated with $O_{n,i}^G$ with respect to the GMM constructed by $O_{m,j}^G$, and select the maximum probability value for the following Gibbs sampling. 
Because there are several overlapping keyframes between two adjacent keyframe groups, some object measurements in these keyframes may be associated with two object landmarks belonging to two different keyframe groups. 
Thus, when executing object association among keyframe groups, the prior probability that two object landmarks should be associated with the same landmark is higher. We enhance the probability by a scale parameter, which is set to 1.5 in our experiments. 

Finally, an object landmark $O_{n,i}^G$ in a keyframe group is assigned a unique object landmark ID $p$ in the global map, \textit{i.e.} $q_{p}=<O_{p}, O^{G}_{n,i}>$. We assume there are $P$ object landmarks in the map at time $t$.

As shown in Fig.\ref{Fig.4.}, the light red box describes the object association process between keyframe groups. Through the object association among keyframe groups, the real object landmarks are finally obtained.

\subsection{Pose Optimization of Object Landmark}\label{3.2}

After object association in the global map, each object landmark has been associated with multiple object measurements. How to determine the pose of this object landmark should be carefully considered. Because the accurate pose of object landmark can help the optimization of SLAM backend, and also can assist to more accurate and robust localization and mapping. In the HDP-SLAM\cite{zhang2019hierarchical}, the pose of first object measurement associated with one object landmark is used as the pose of this landmark, which is obviously not reasonable.

In this study, we take all poses of object measurements associated with one object landmark into account. For each object measurement, the distance and angle between it and other object measurements are computed and summed up. The pose of one object measurement that has the smallest distance and angle is used as the pose of object landmark. The angle difference ($\theta_{k,l}$) and the distance difference ($\varphi_{k,l}$) of the $k$th object measurement to the $l$th object measurement associated with the same object landmark are computed. Because the scales of distance and angle are different, we need to normalize them. We set the maximal angle difference $A$ and maximal distance difference $B$, then the normalized average angle difference and average distance difference are computed as:

\begin{equation}
	\bar{\theta}_{k,l} = 
	\begin{cases}
		1 & \text{if } \theta_{k,l}>A, \\
		\theta_{k,l}/A & \text{otherwise}.
	\end{cases}
\end{equation}
\begin{equation}	
	\bar{\varphi}_{k,l} = 
	\begin{cases}
		1 & \text{if } \varphi_{k,l}>B, \\
		\varphi_{k,l}/B & \text{otherwise}.
	\end{cases}	
\end{equation}

The average pose difference of the $k$th object measurement associated with the object landmark $O_p$ is then computed as:

\begin{equation}
f_{O_p}(L_k) =\alpha \times \frac{ \sum_{l\in Z\backslash k}{\bar{\theta}_{k,l}} }{|Z|-1} + \beta \times \frac{ \sum_{l\in Z\backslash k}\bar{\varphi}_{k,l} }{|Z|-1},
\label{equ.10}
\end{equation}
where $Z$ means all object measurements associated with $O_p$, $Z\backslash k$ denotes all object measurements except $L_k$. $\alpha$ and $\beta$ are the weights for angle and position differences, respectively. Through the analysis of experimental results, we set them to 0.4 and 0.6 for the best effect, respectively. Then we sort all pose difference and set the pose of object landmark to the pose of the object measurement who has the minimal pose difference.

\section{Experiment results}

\begin{table}[]
\caption{ The table compares the average time cost of each main process. }
\begin{center}
\setlength{\tabcolsep}{4.5mm}{
\begin{tabular}{|c|c|c|lllllll}
\cline{1-3}
\multirow{2}{*}{Process   Name} & \multicolumn{2}{c|}{Time Consuming(s)} &  &  &  &  &  &  &  \\ \cline{2-3}
                                & Our                & HDP-SLAM\cite{zhang2019hierarchical}          &  &  &  &  &  &  &  \\ \cline{1-3}
MOT          & 0.052490           & -          &  &  &  &  &  &  &  \\ \cline{1-3}
Object Detection         & -           & 0.022228          &  &  &  &  &  &  &  \\ \cline{1-3}
Pose Prediction                 & 0.012301           & 0.012902          &  &  &  &  &  &  &  \\ \cline{1-3}
Object Association              & 0.016704           & 0.013896          &  &  &  &  &  &  &  \\ \cline{1-3}
Object Optimization             & 0.013042           & 0.013359          &  &  &  &  &  &  &  \\ \cline{1-3}
ORB-SLAM2 Tracking              & 0.023670           & 0.025137          &  &  &  &  &  &  &  \\ \cline{1-3}
\end{tabular}}
\end{center}
\label{tab5}  
\end{table}

\begin{table*}[ht]
\caption{ Comparison experimental results of simulated hospital sequences.}
 \begin{center}
\setlength{\tabcolsep}{2.5mm}{
\begin{tabular}{|c|l|c|l|c|l|c|c|l|c|c|c|l|c|}
\hline
\multicolumn{2}{|c|}{\multirow{3}{*}{sequence number}} & \multicolumn{5}{c|}{Absolute Trajectory RMSE(m)}                                                                   & \multicolumn{3}{c|}{Association accuracy(\%)}                                & \multicolumn{4}{c|}{Object number}                                                                   \\ \cline{3-14} 
\multicolumn{2}{|c|}{}                                 & \multicolumn{2}{c|}{\multirow{2}{*}{ORB-SLAM2\cite{mur2017orb}}} & \multicolumn{2}{c|}{\multirow{2}{*}{HDP-SLAM\cite{zhang2019hierarchical}}} & \multirow{2}{*}{Ours}      & \multicolumn{2}{c|}{\multirow{2}{*}{HDP-SLAM\cite{zhang2019hierarchical}}} & \multirow{2}{*}{Ours}       & \multirow{2}{*}{Object\_GT} & \multicolumn{2}{c|}{\multirow{2}{*}{HDP-SLAM\cite{zhang2019hierarchical}}} & \multirow{2}{*}{Ours} \\
\multicolumn{2}{|c|}{}                                 & \multicolumn{2}{c|}{}                           & \multicolumn{2}{c|}{}                          &                            & \multicolumn{2}{c|}{}                          &                             &                             & \multicolumn{2}{c|}{}                          &                       \\ \hline
\multicolumn{2}{|c|}{aisle\_slow\_1}                   & \multicolumn{2}{c|}{0.113}                  & \multicolumn{2}{c|}{0.086}                  & \textbf{0.058}          & \multicolumn{2}{c|}{73.03}                     & \textbf{81.08}              & 6                           & \multicolumn{2}{c|}{8}                         & \textbf{6}                     \\ \hline
\multicolumn{2}{|c|}{aisle\_slow\_2}                   & \multicolumn{2}{c|}{0.132}                  & \multicolumn{2}{c|}{0.088}                  & \textbf{0.054}          & \multicolumn{2}{c|}{74.11}                     & \textbf{89.53}              & 8                           & \multicolumn{2}{c|}{12}                        & \textbf{8}                     \\ \hline
\multicolumn{2}{|c|}{aisle\_slow\_3}                   & \multicolumn{2}{c|}{0.066}                   & \multicolumn{2}{c|}{0.044}                  & \textbf{0.034}          & \multicolumn{2}{c|}{62.02}                     & \textbf{78.84}              & 8                           & \multicolumn{2}{c|}{14}                        & \textbf{8}                     \\ \hline
\multicolumn{2}{|c|}{aisle\_quick\_1}                  & \multicolumn{2}{c|}{0.233}                  & \multicolumn{2}{c|}{0.204}                 & \textbf{0.146}         & \multicolumn{2}{c|}{75.47}                     & \textbf{96.87}              & 6                           & \multicolumn{2}{c|}{7}                         & \textbf{6}                     \\ \hline
\multicolumn{2}{|c|}{aisle\_quick\_2}                  & \multicolumn{2}{c|}{0.211}                  & \multicolumn{2}{c|}{0.127}                 & \textbf{0.094}          & \multicolumn{2}{c|}{51.32}                     & \textbf{92.59}              & 8                           & \multicolumn{2}{c|}{12}                        & \textbf{8}                     \\ \hline
\multicolumn{2}{|c|}{aisle\_quick\_3}                  & \multicolumn{2}{c|}{0.198}                  & \multicolumn{2}{c|}{0.100}                 & \textbf{0.074}          & \multicolumn{2}{c|}{51.19}                     & \textbf{77.42}              & 8                           & \multicolumn{2}{c|}{18}                        & \textbf{9}                     \\ \hline
\multicolumn{2}{|c|}{office\_desk\_slow}               & \multicolumn{2}{c|}{0.146}                  & \multicolumn{2}{c|}{0.064}                  & \textbf{0.045}          & \multicolumn{2}{c|}{61.11}                     & \textbf{97.53}              & 5                           & \multicolumn{2}{c|}{9}                         & \textbf{5}                     \\ \hline
\end{tabular}}
\end{center}
\label{tab1}  
\end{table*}

To evaluate the proposed method comprehensively, we carry out two kinds of experiments on three kinds of scenes. One verifies the improvements of object association by the proposed hierarchical grouping and association strategy. Another verifies the accuracy of the trajectory errors by comparing with SOTA semantic/visual SLAM systems. Our experiments are executed on seven sequences of simulated hospital environments, three sequences of real hospital environments and six sequences in the KITTI\cite{geiger2013vision} dataset. Our SLAM system is run on a 3.6GHz eight-core Linux system with the RTX2070 graphics card. All experiments run in real time. The average computation time of each main process is listed in Table \ref{tab5}, and the average time cost of inserting a keyframe is 0.198462 seconds.

There are many semantic SLAM systems, such as CubeSLAM\cite{yang2019cubeslam}, VPS-SLAM\cite{bavle2020vps}, AVP-SLAM\cite{qin2020avp}, and TextSLAM\cite{li2020textslam}. However, it is unfortunate that they are either not open source or not suitable for our current environment, so that we cannot compare our method with them from the aspect of object association. We compare our method with these SOTA semantic SLAM systems in terms of trajectory errors on the KITTI dataset according to their results in their papers.
To evaluate the performance of object association, we compare our methods with the HDP-SLAM\cite{zhang2019hierarchical} in all three kinds of scenes.

\begin{figure*} [ht]

\centering 
\subfigure[Object association and object poses in aisle\_slow\_1.]{ 
  
\includegraphics[width=3in, height=2.25in]{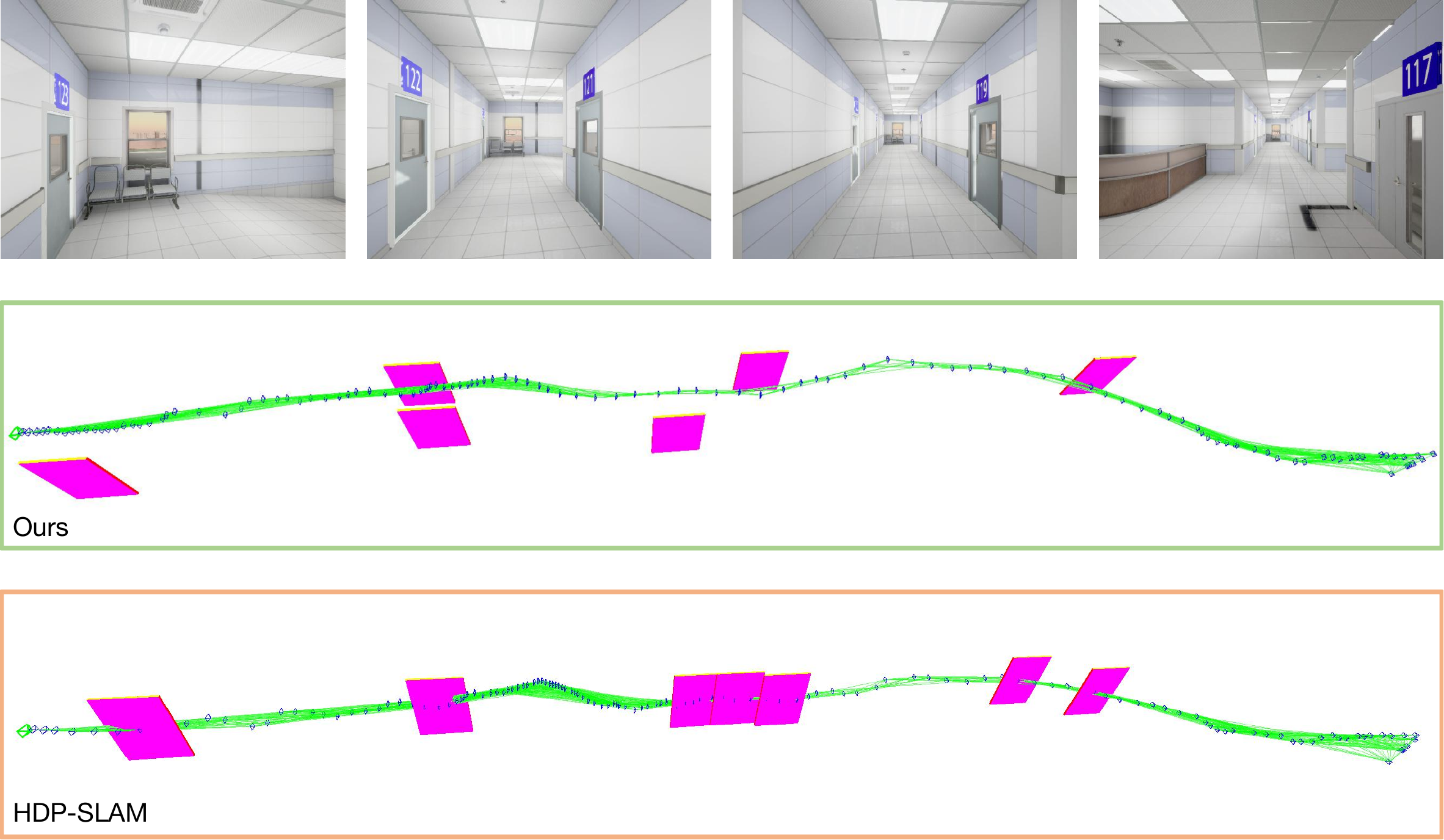}} \qquad
\centering 
\subfigure[Object association and object poses in aisle\_quick\_2.]{ 
  
\includegraphics[width=3in, height=2.25in]{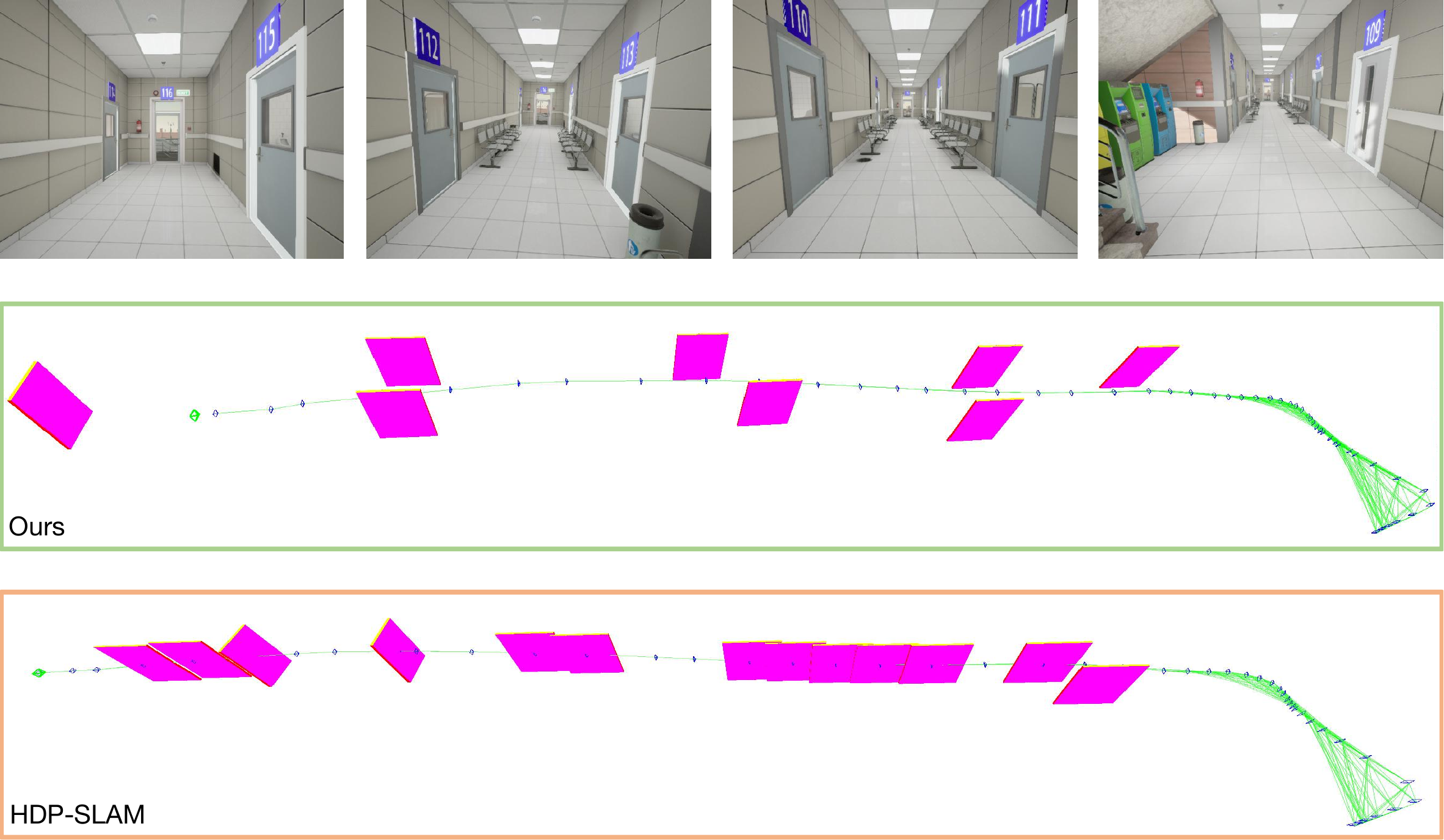}}

\centering 
\subfigure[Trajectories in the aisle\_slow\_1.]{ 
  
\includegraphics[width=1.45in, height=1.45in, trim=10 10 10 10, clip]{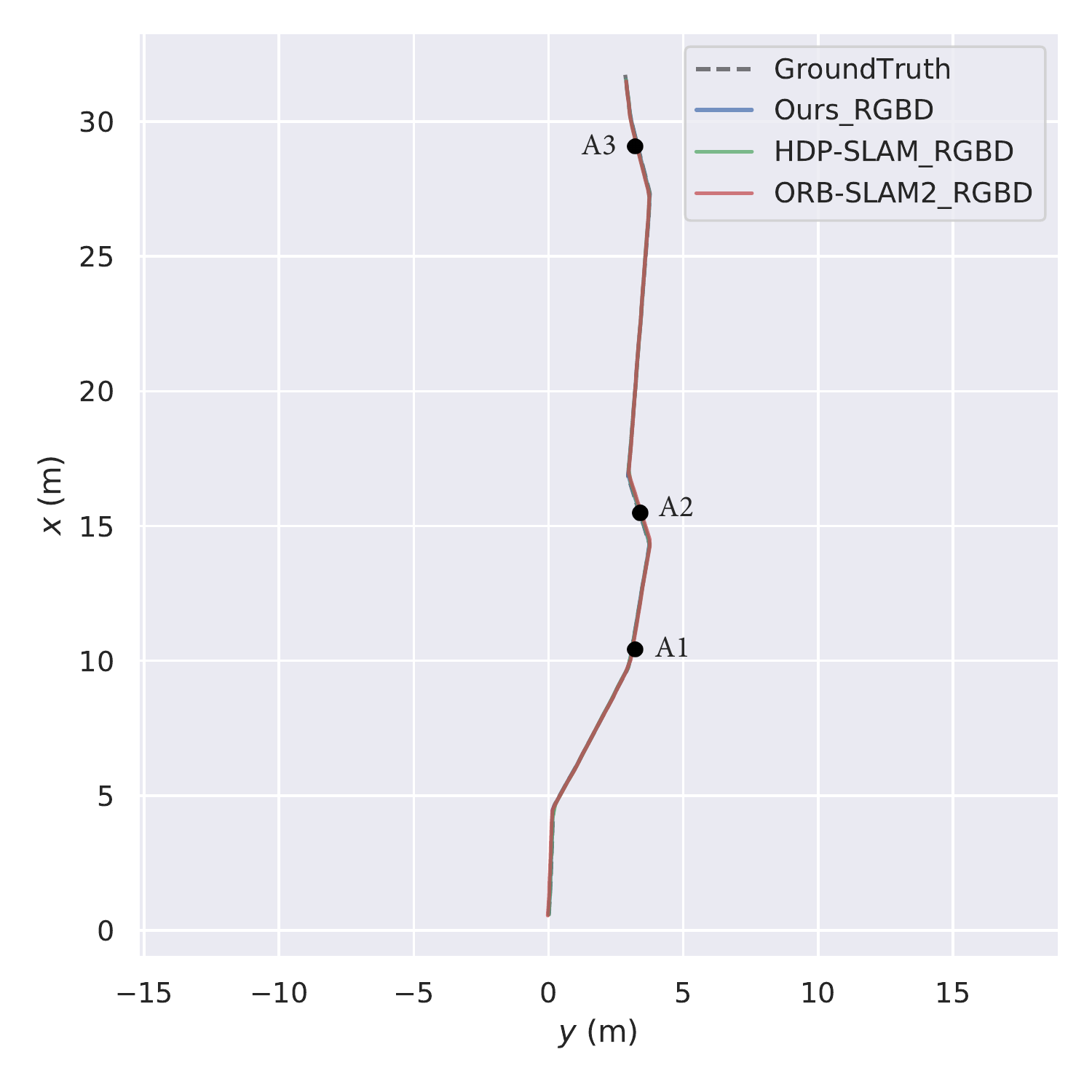}} \qquad
\centering 
\subfigure[Trajectories near point A1.]{ 
  
\includegraphics[width=1.45in, height=1.45in, trim=10 10 10 10, clip]{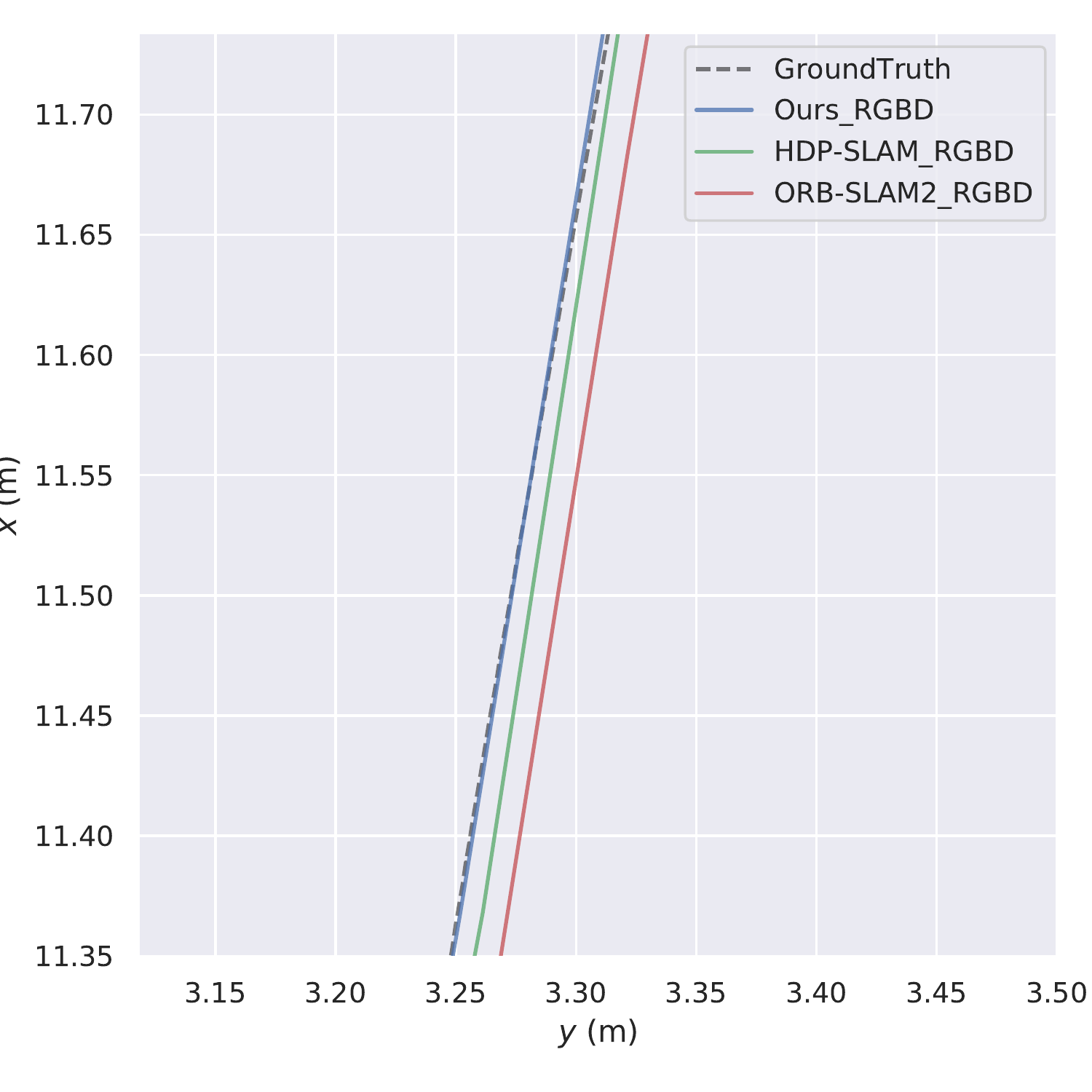}} \qquad
\centering 
\subfigure[Trajectories near point A2.]{ 
  
\includegraphics[width=1.45in, height=1.45in, trim=10 10 10 10, clip]{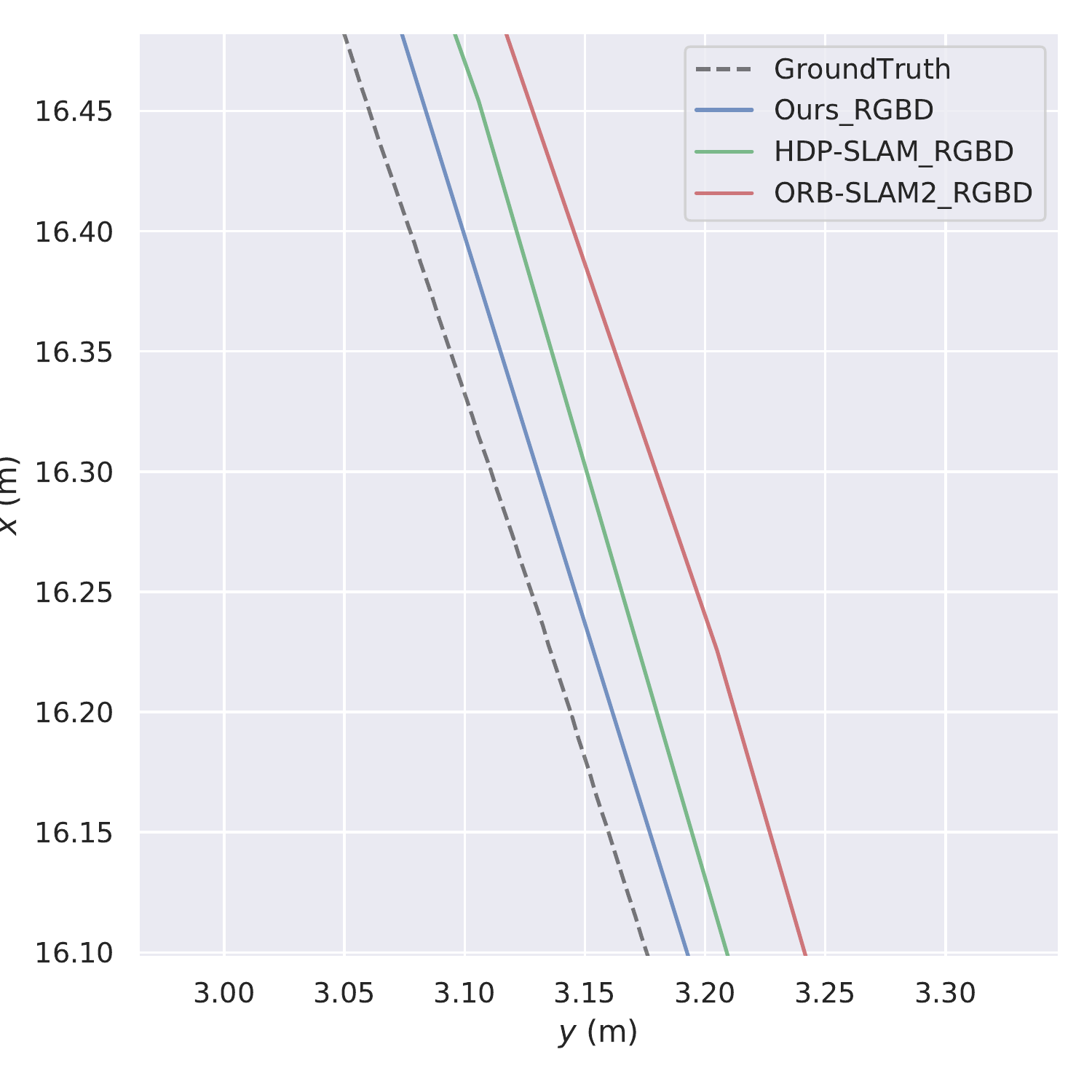}} \qquad
\centering 
\subfigure[Trajectories near point A3.]{ 
  
\includegraphics[width=1.45in, height=1.45in, trim=10 10 10 10, clip]{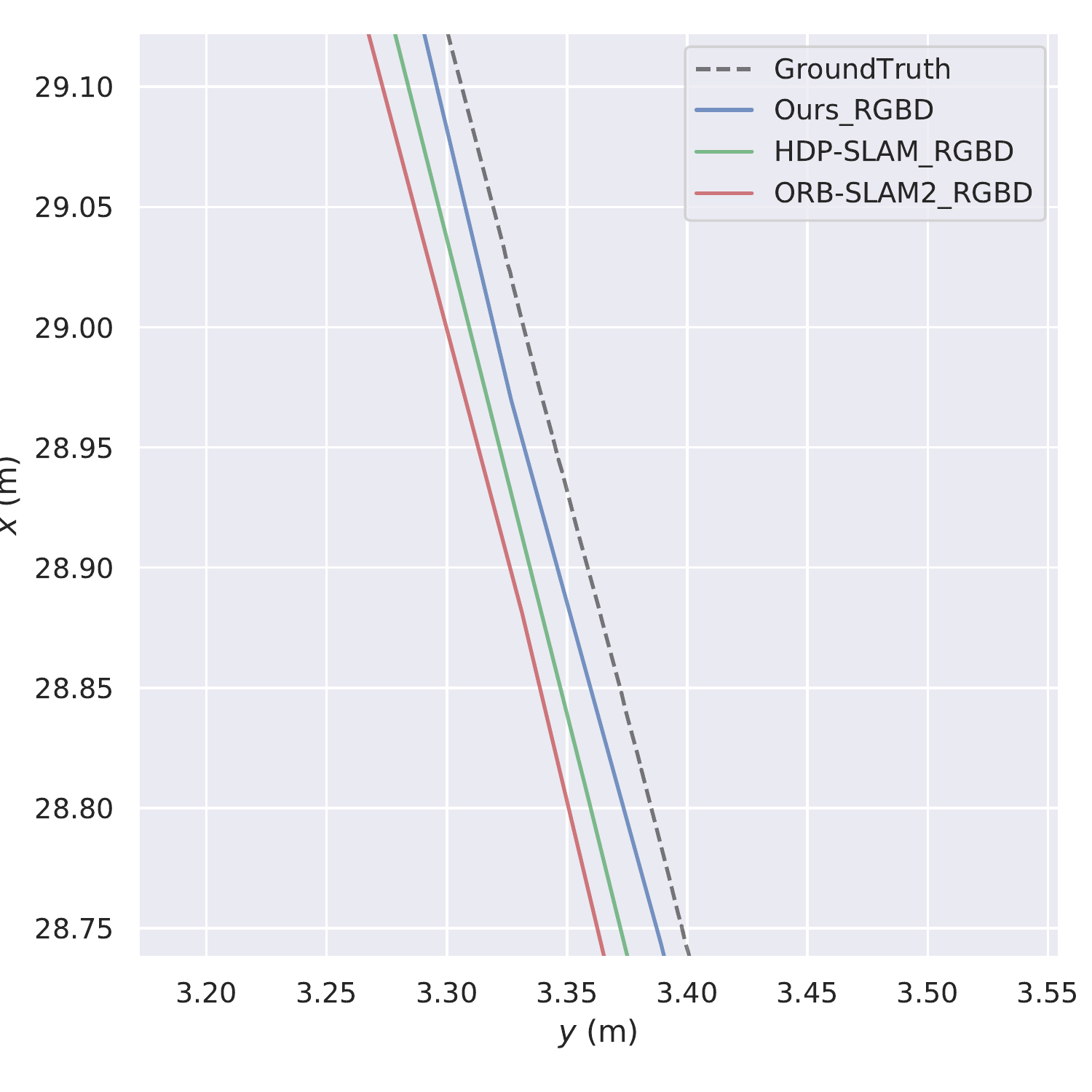}}

\centering 
\subfigure[Trajectories in the aisle\_quick\_2.]{ 
  
\includegraphics[width=1.45in, height=1.45in, trim=10 10 10 10, clip]{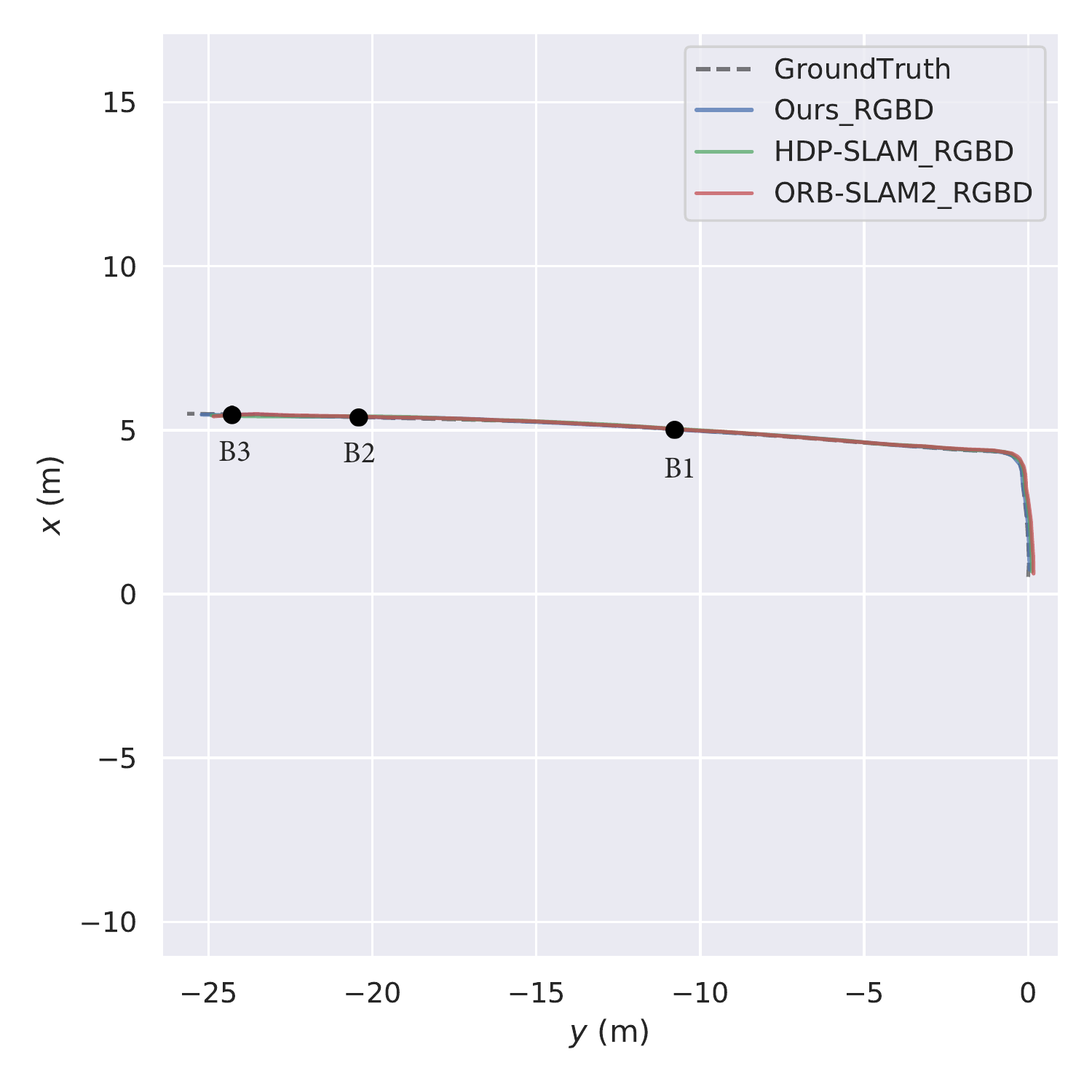}} \qquad
\centering 
\subfigure[Trajectories near point B1.]{ 
  
\includegraphics[width=1.45in, height=1.45in, trim=10 10 10 10, clip]{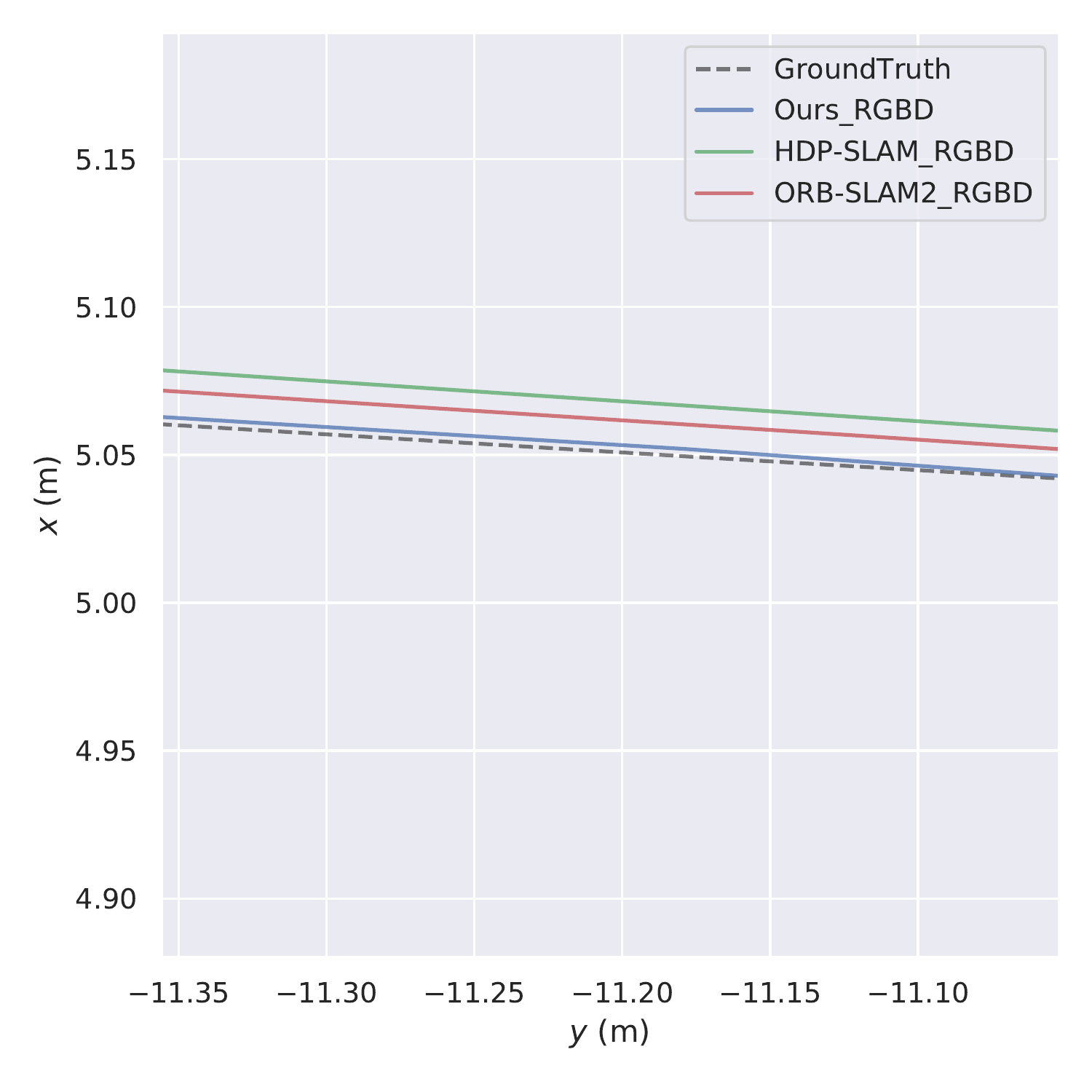}} \qquad
\centering 
\subfigure[Trajectories near point B2.]{ 
  
\includegraphics[width=1.45in, height=1.45in, trim=10 10 10 10, clip]{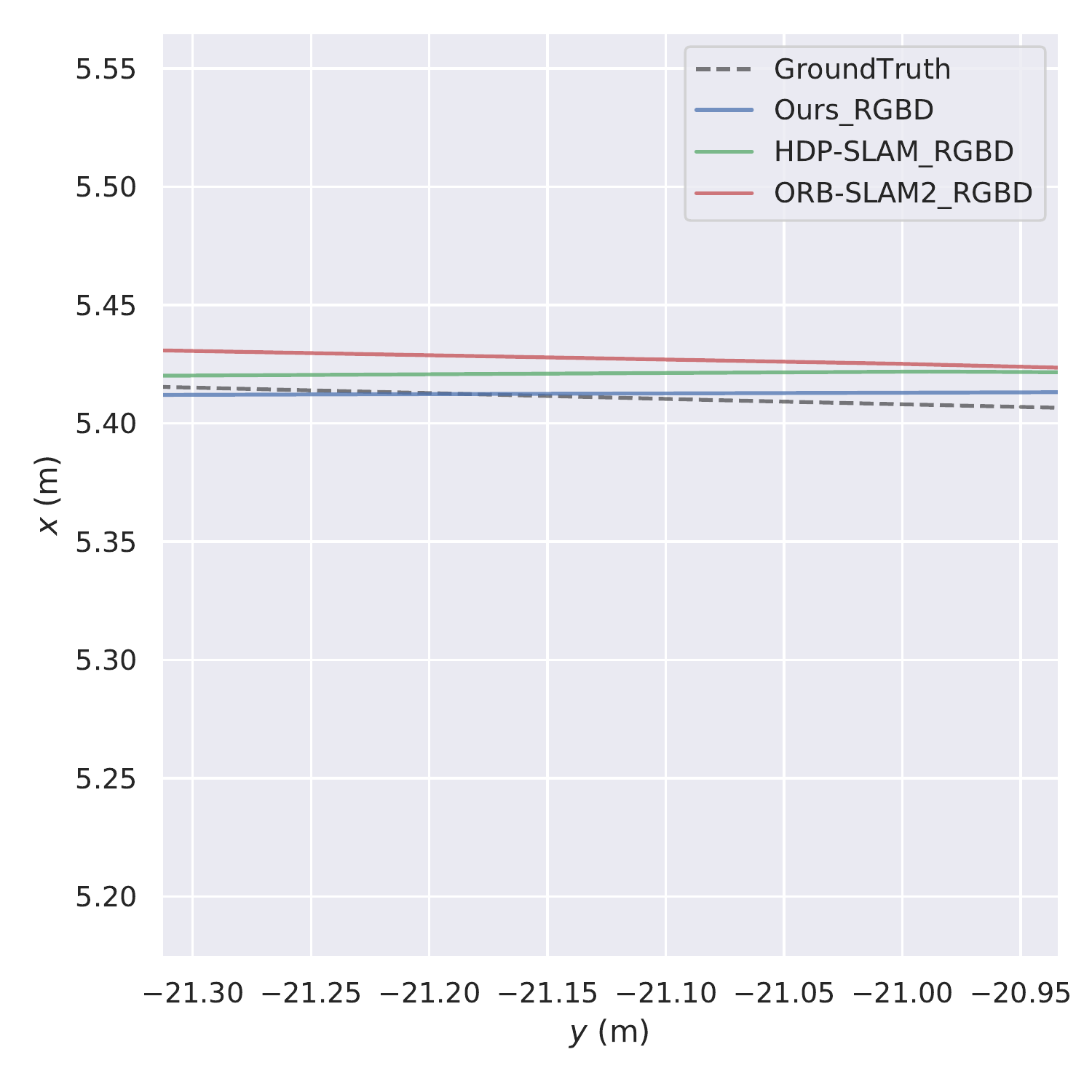}} \qquad
\centering  
\subfigure[Trajectories near point B3.]{ 
  
\includegraphics[width=1.45in, height=1.45in, trim=10 10 10 10, clip]{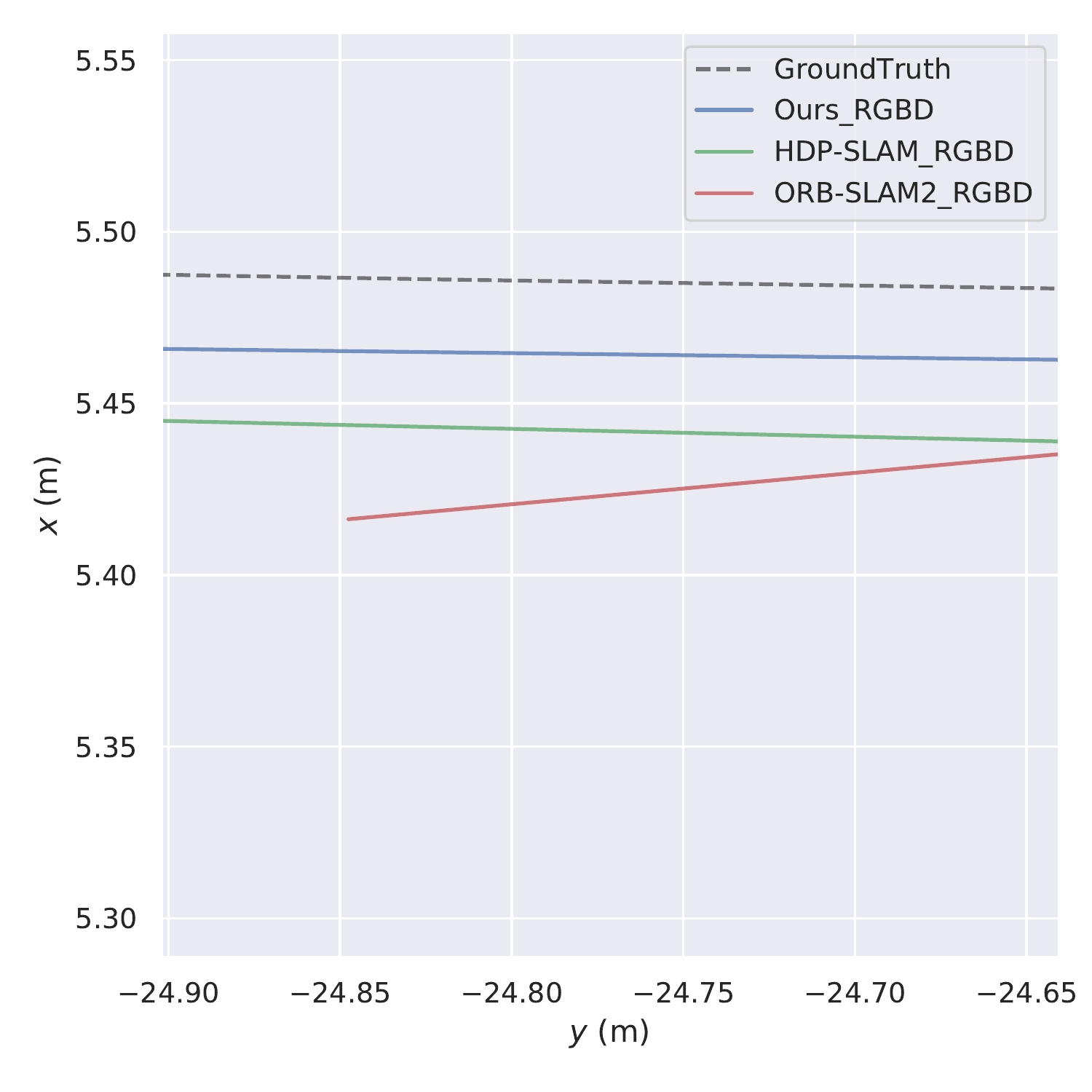}}

\caption{ Comparison of object poses and trajectories in simulated hospital sequences. In Figures (a) and (b), the first row shows the environment of sequences, the second row shows the trajectories and object maps which is obtained using the proposed method, and the third row shows the trajectories and object maps using HDP-SLAM. In the second and third rows, the red cuboid represents the doors. In Figures (c-f), comparison of trajectories in the aisle\_slow\_1. In Figures (g-j), comparison of trajectories in the aisle\_quick\_2.} 
\label{Fig.5.} 
\end{figure*}

\begin{figure*} 
\vspace*{-10pt}

\centering 
\subfigure[Trajectory comparison results of the KITTI-03.]{ 
  
\includegraphics[width=1.3in, height=1.3in, trim=10 10 10 10, clip]{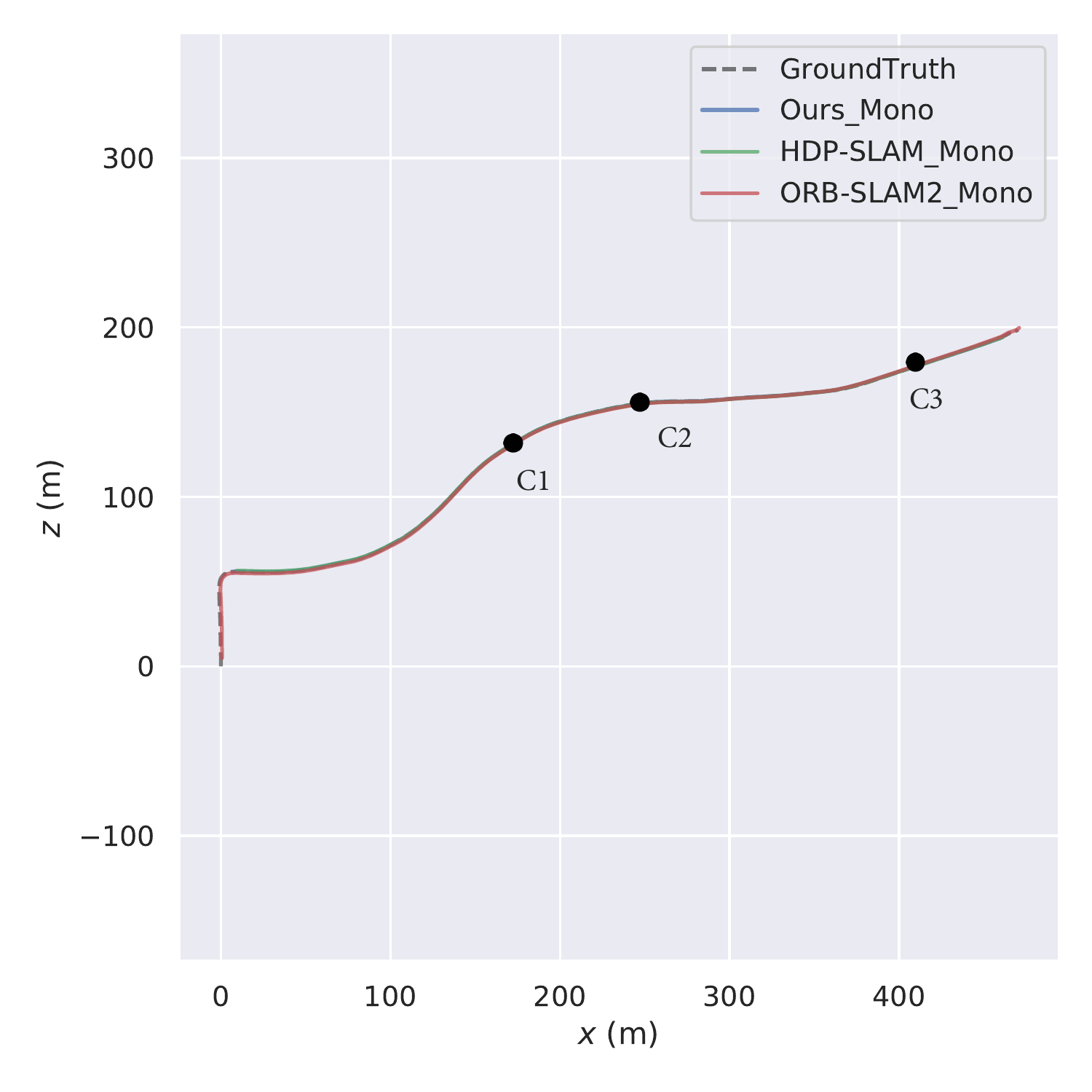}} \qquad
\centering 
\subfigure[Comparison of trajectories near point C1.]{ 
  
\includegraphics[width=1.3in, height=1.3in, trim=10 10 10 10, clip]{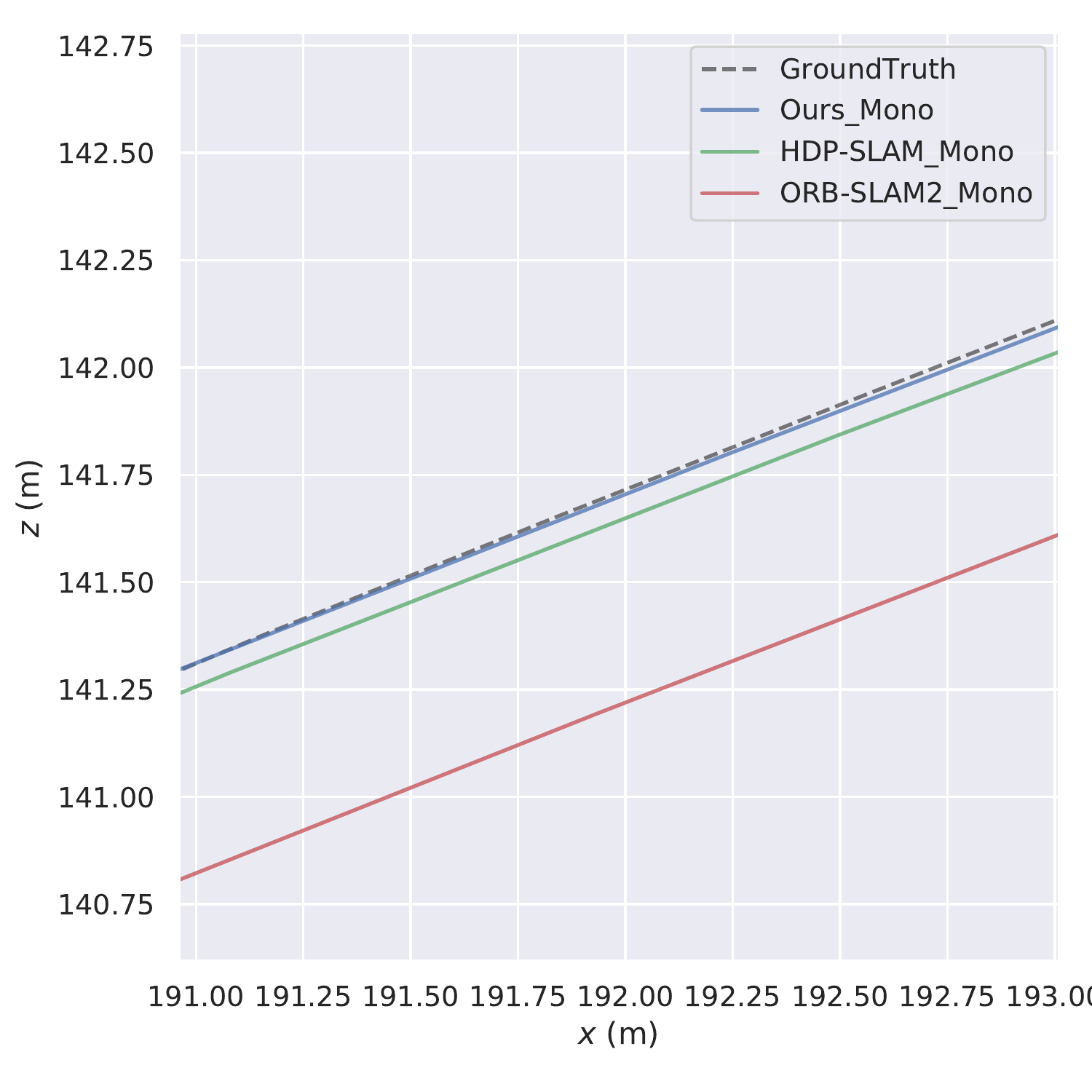}} \qquad
\centering 
\subfigure[Comparison of trajectories near point C2.]{ 
  
\includegraphics[width=1.3in, height=1.3in, trim=10 10 10 10, clip]{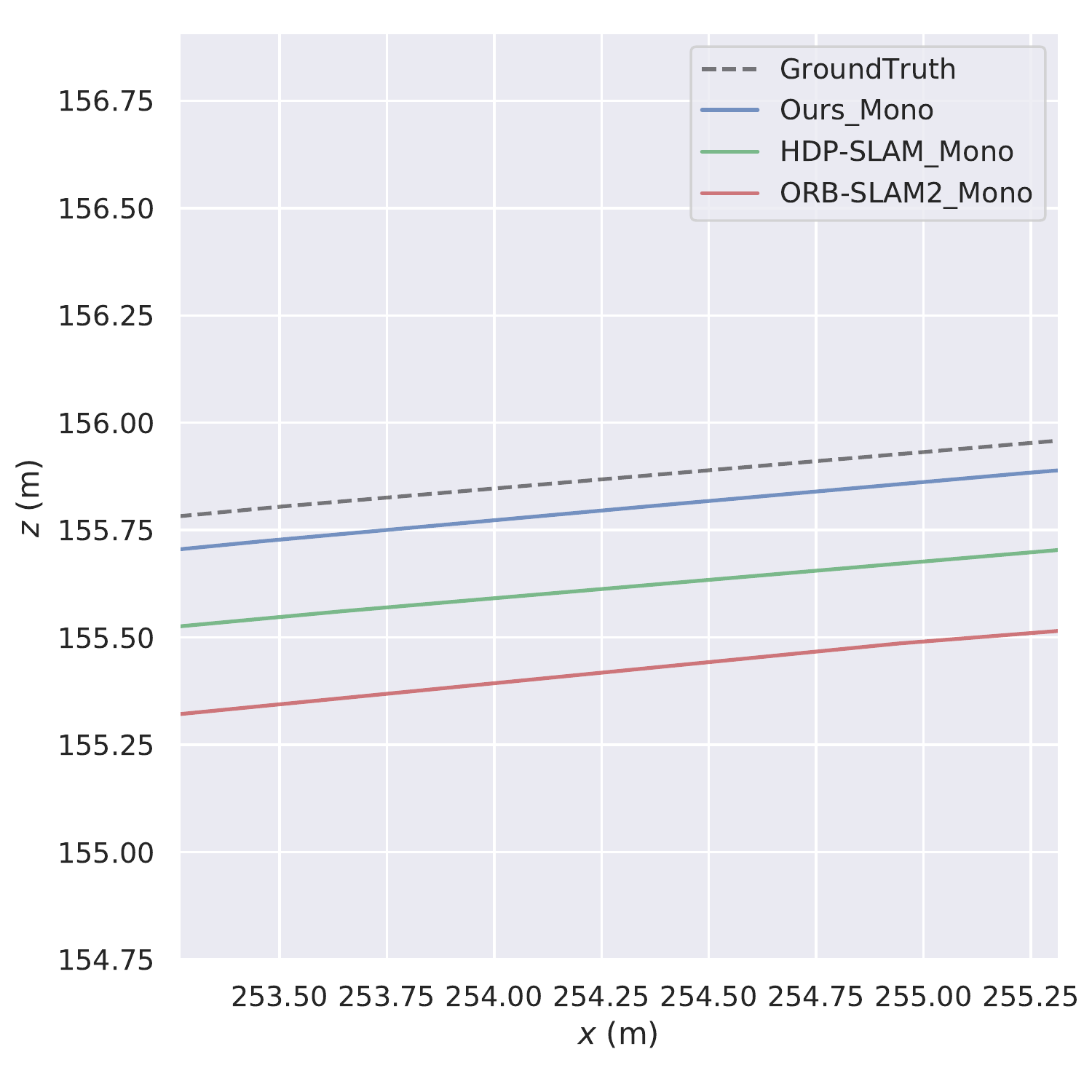}} \qquad
\centering 
\subfigure[Comparison of trajectories near point C3.]{ 
  
\includegraphics[width=1.3in, height=1.3in, trim=10 10 10 10, clip]{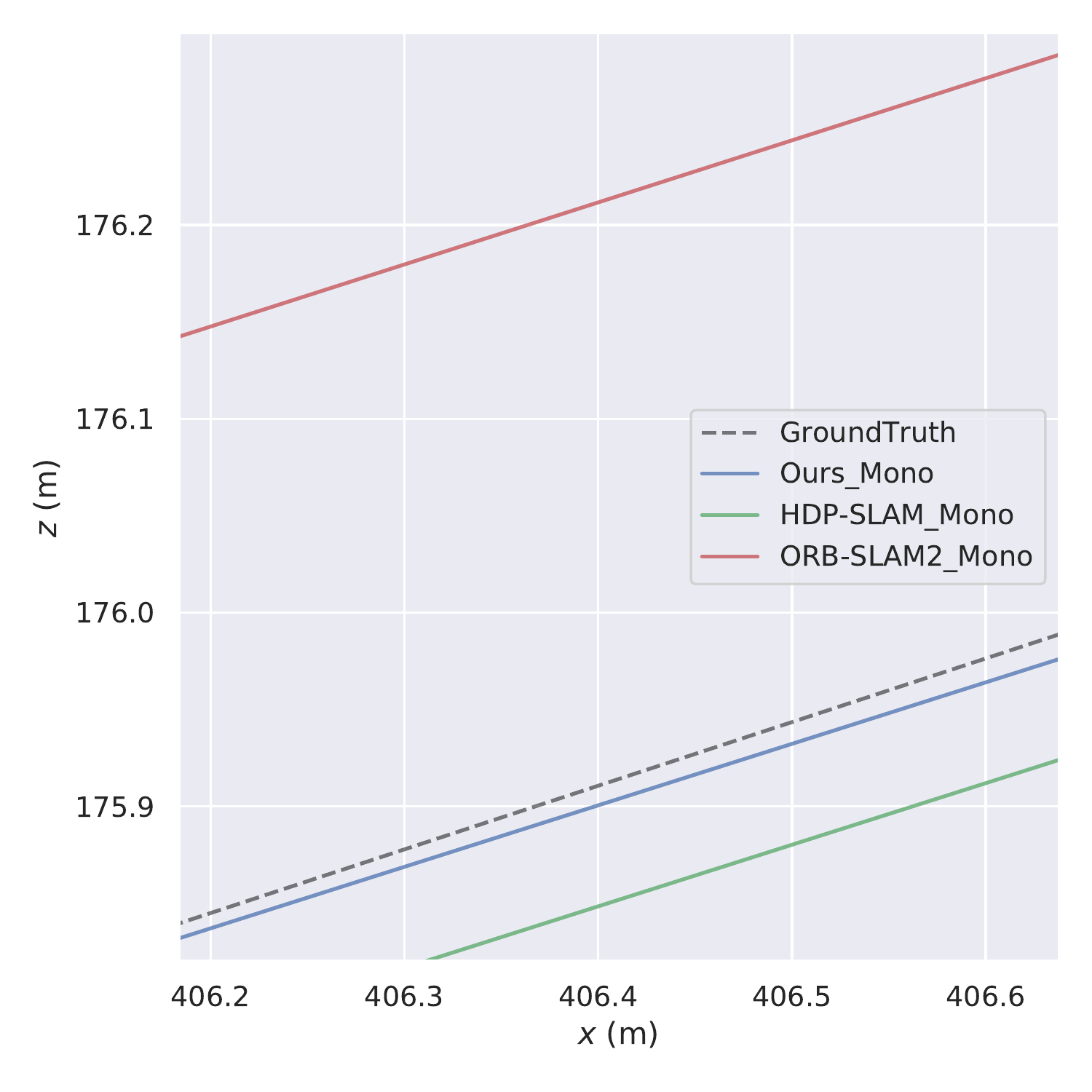}}

\centering 
\subfigure[Trajectory comparison results of the KITTI-05.]{ 
  
\includegraphics[width=1.3in, height=1.3in, trim=10 10 10 10, clip]{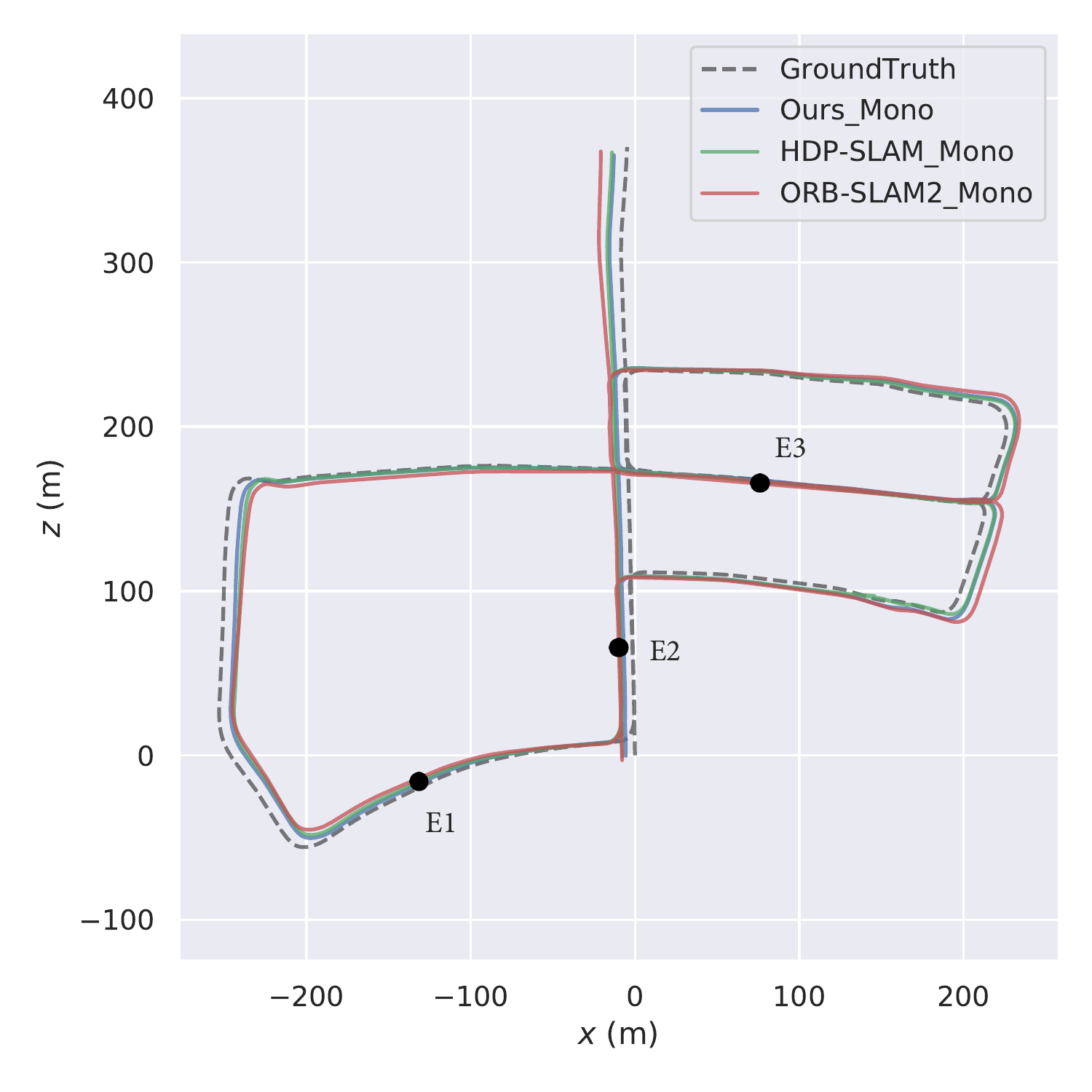}} \qquad
\centering 
\subfigure[Comparison of trajectories near point E1.]{ 
  
\includegraphics[width=1.3in, height=1.3in, trim=10 10 10 10, clip]{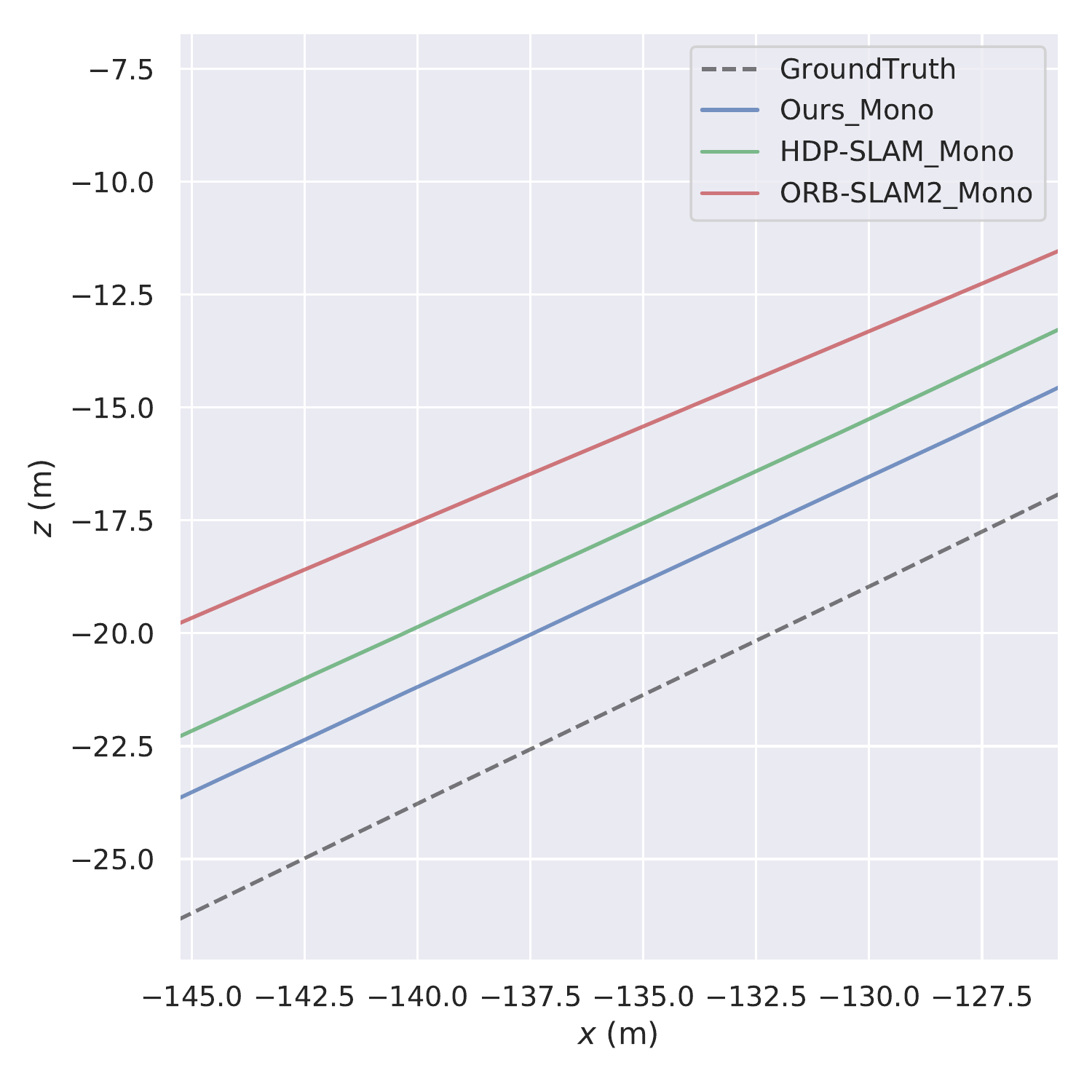}} \qquad
\centering 
\subfigure[Comparison of trajectories near point E2.]{ 
  
\includegraphics[width=1.3in, height=1.3in, trim=10 10 10 10, clip]{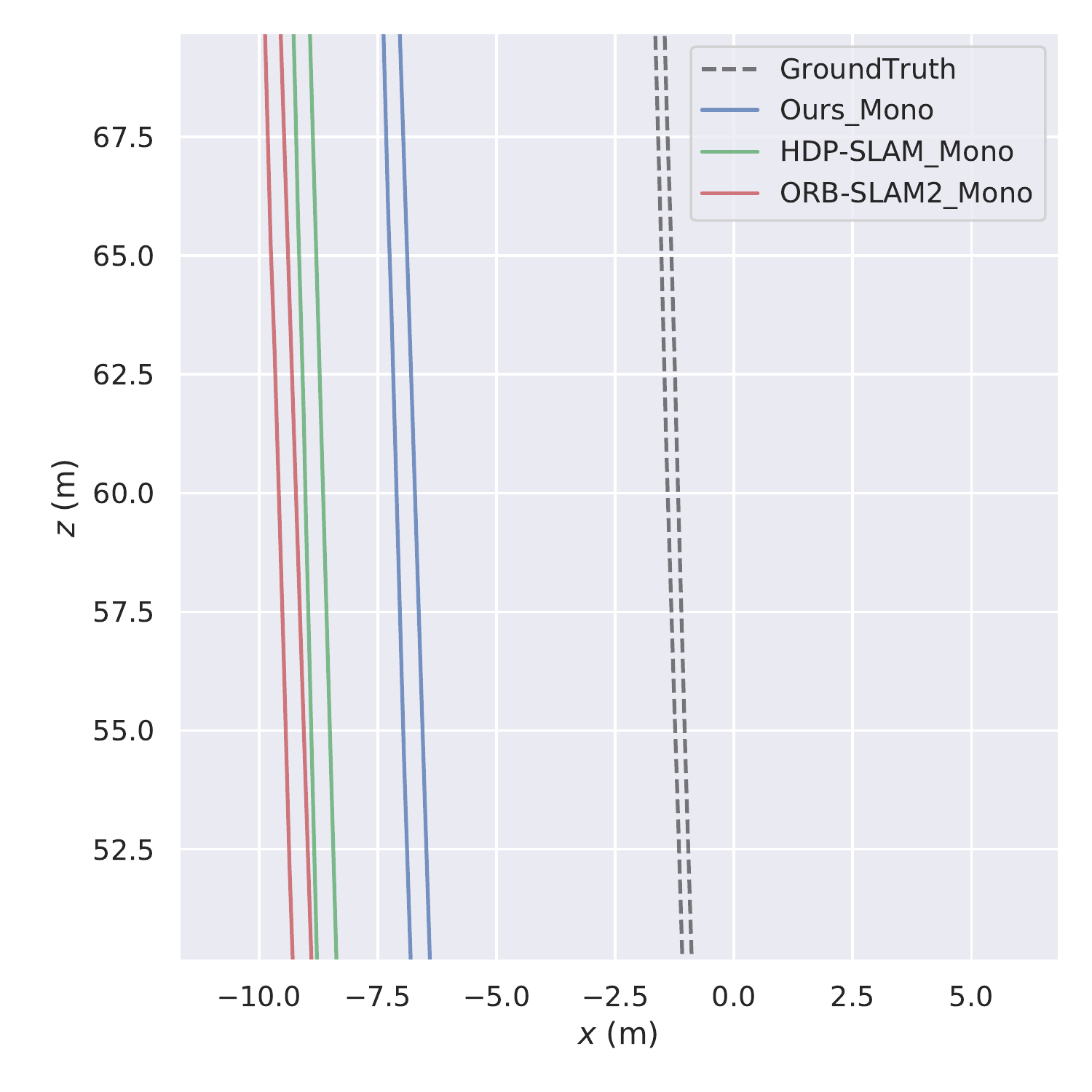}} \qquad
\centering  
\subfigure[Comparison of trajectories near point E3.]{ 
  
\includegraphics[width=1.3in, height=1.3in, trim=10 10 10 10, clip]{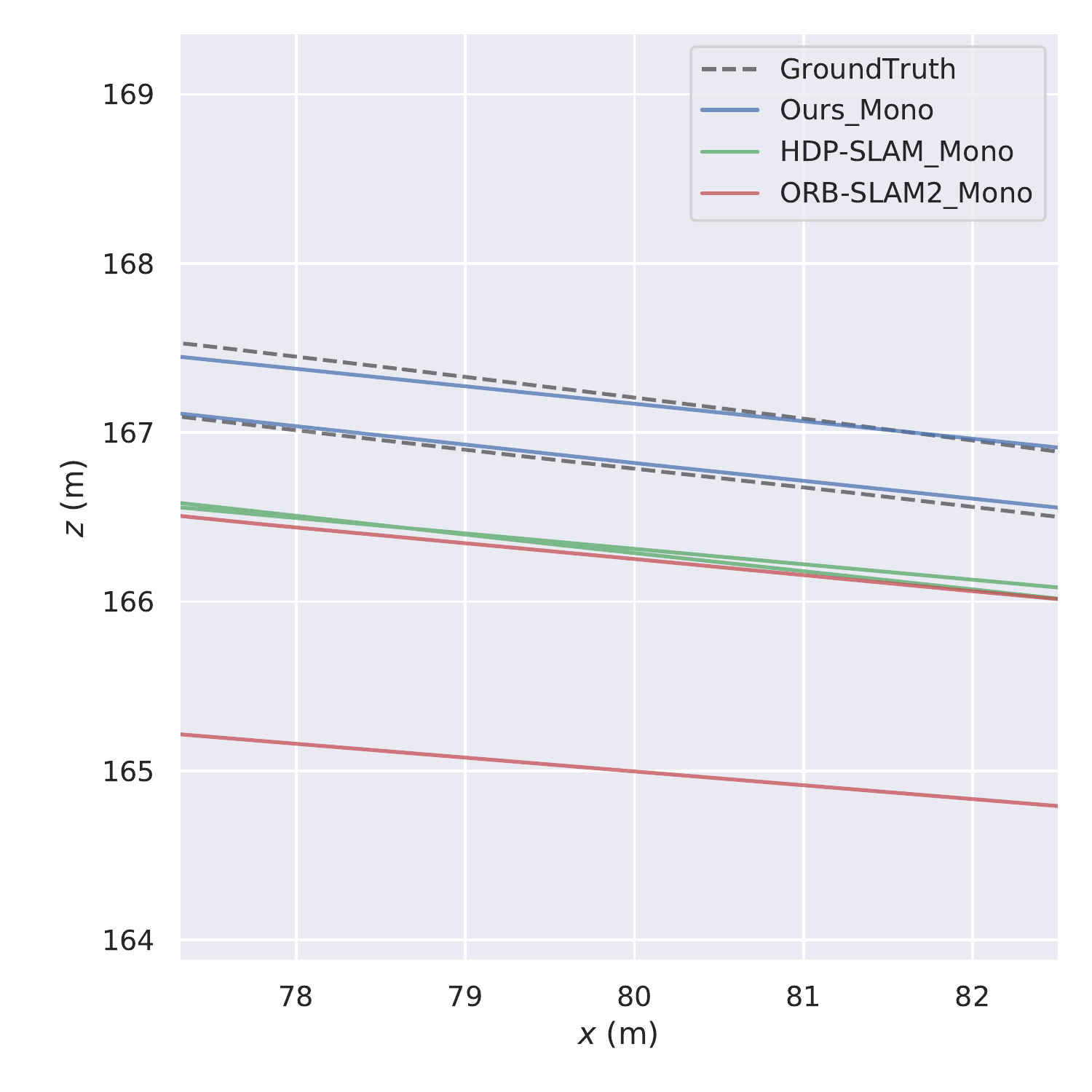}}

\centering 
\subfigure[Trajectory comparison results of the KITTI-07.]{ 
  
\includegraphics[width=1.3in, height=1.3in, trim=10 10 10 10, clip]{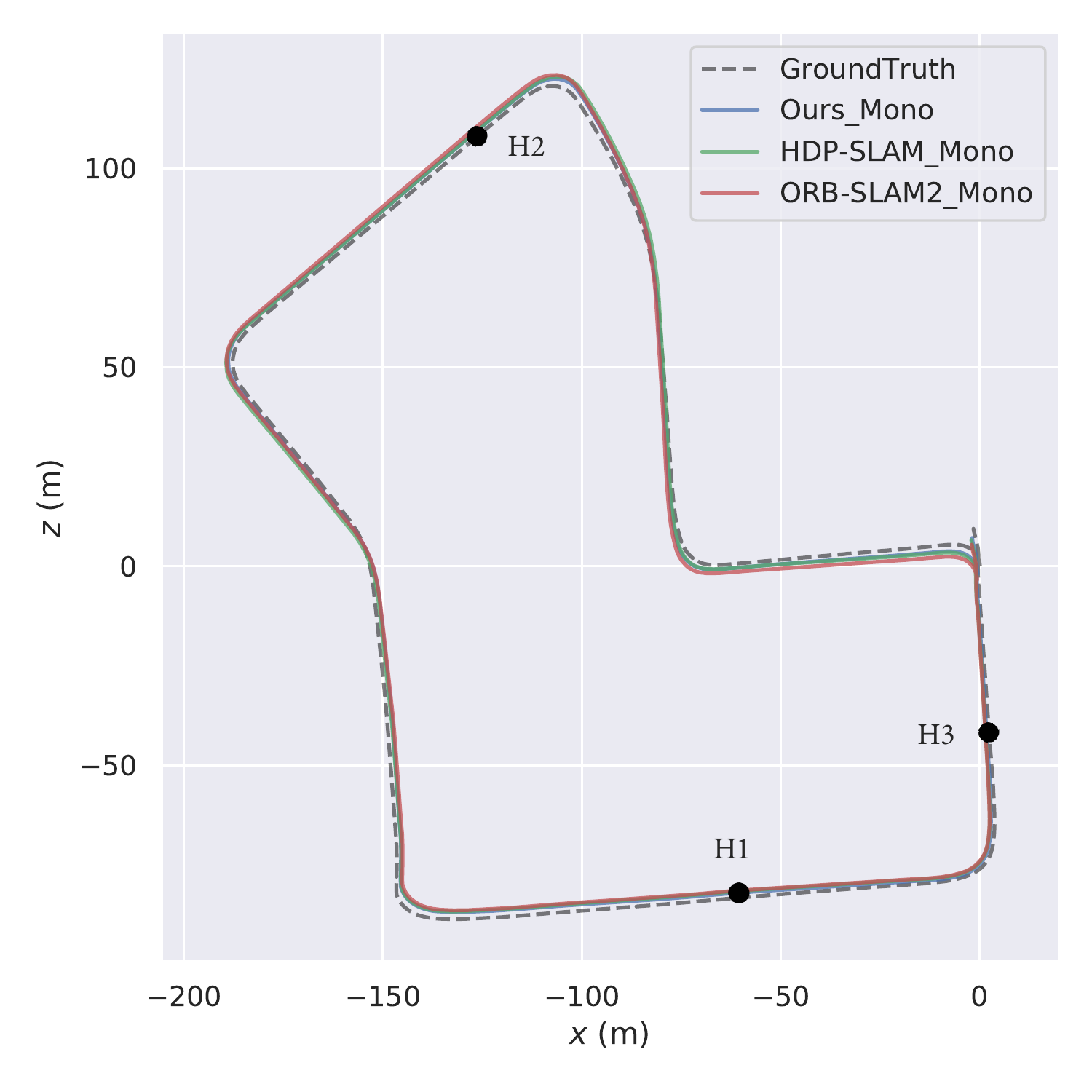}} \qquad
\centering 
\subfigure[Comparison of trajectories near point H1.]{ 
  
\includegraphics[width=1.3in, height=1.3in, trim=10 10 10 10, clip]{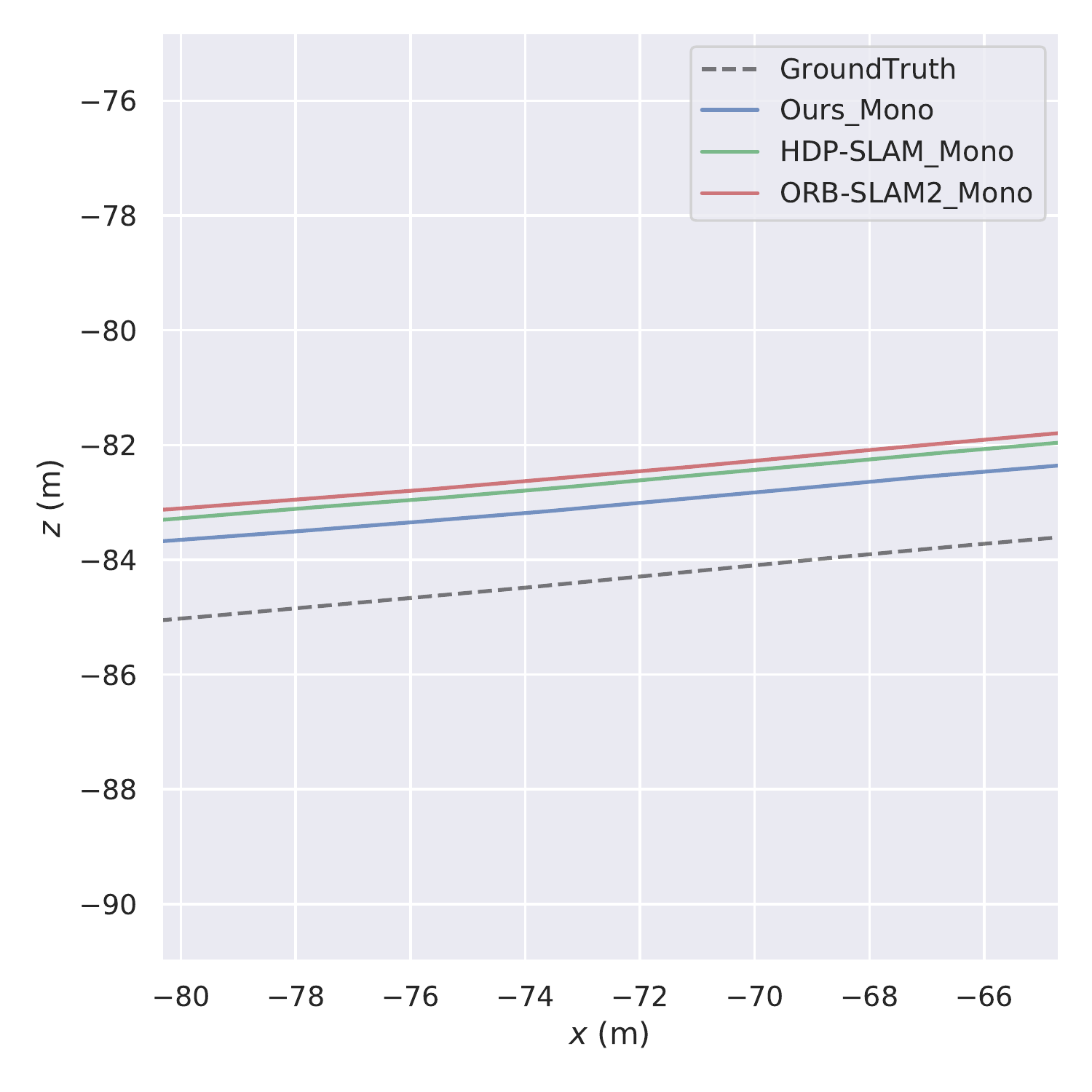}} \qquad
\centering 
\subfigure[Comparison of trajectories near point H2.]{ 
  
\includegraphics[width=1.3in, height=1.3in, trim=10 10 10 10, clip]{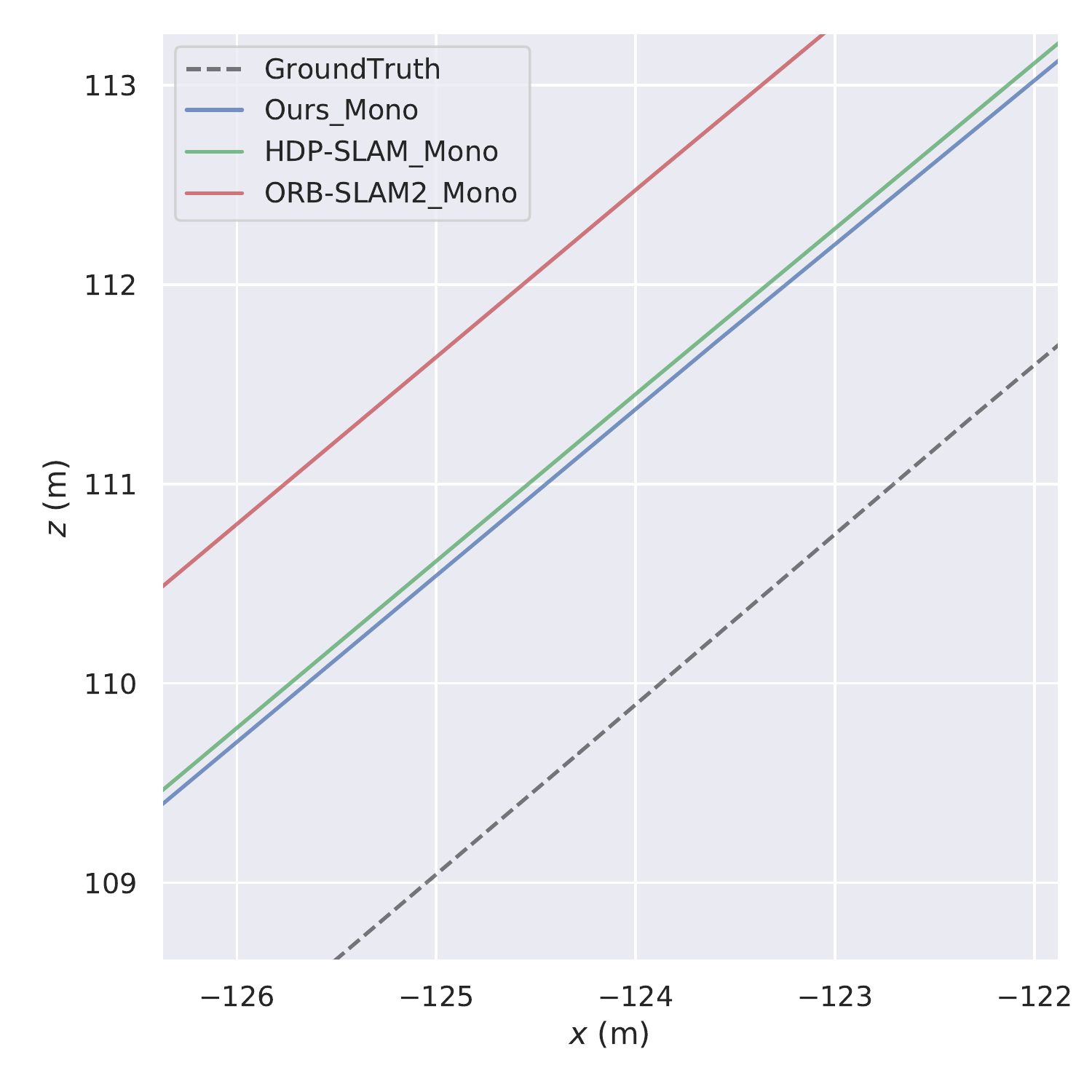}} \qquad
\centering  
\subfigure[Comparison of trajectories near point H3.]{ 
  
\includegraphics[width=1.3in, height=1.3in, trim=10 10 10 10, clip]{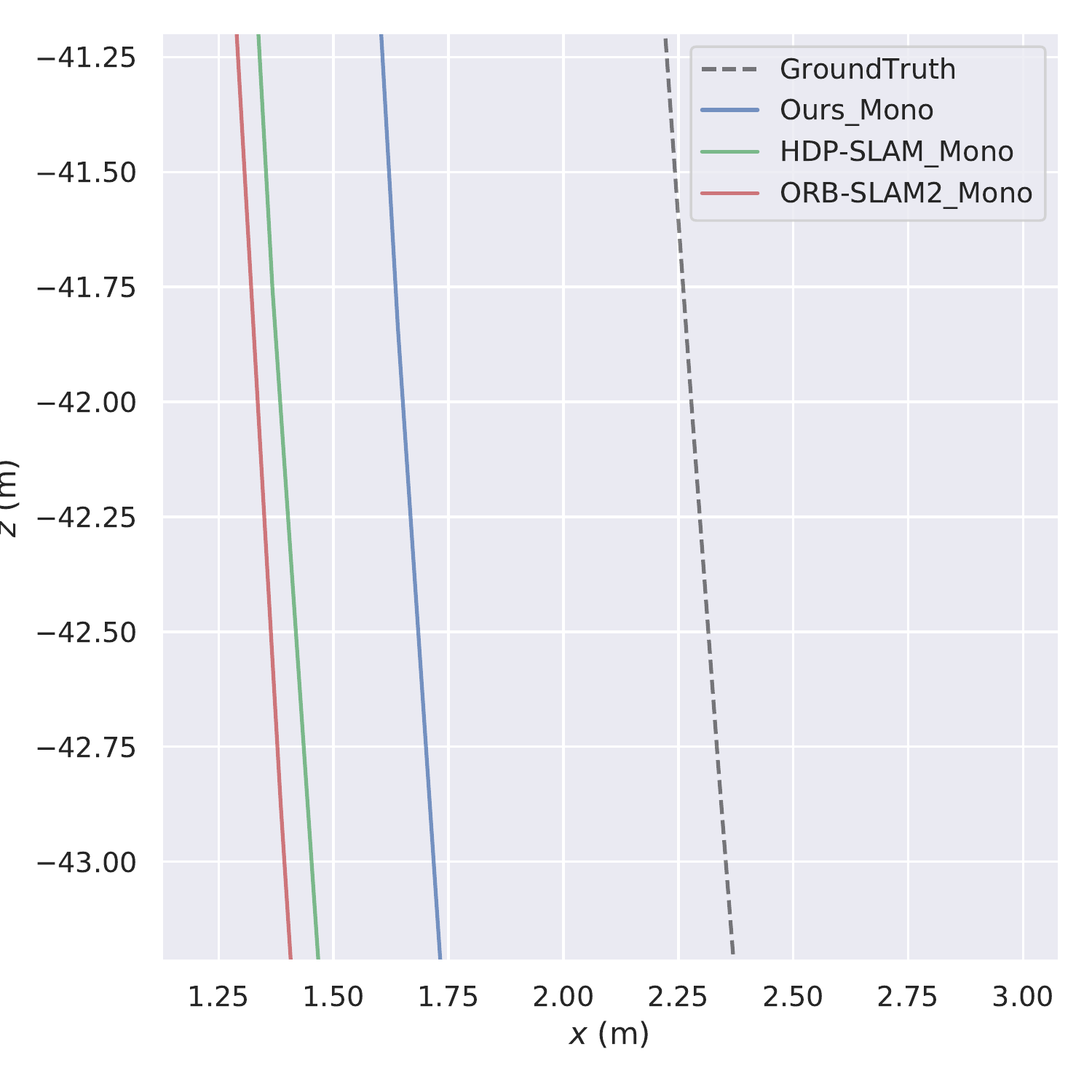}}

\centering 
\subfigure[Trajectory comparison results of the KITTI-10.]{ 
  
\includegraphics[width=1.3in, height=1.3in, trim=10 10 10 10, clip]{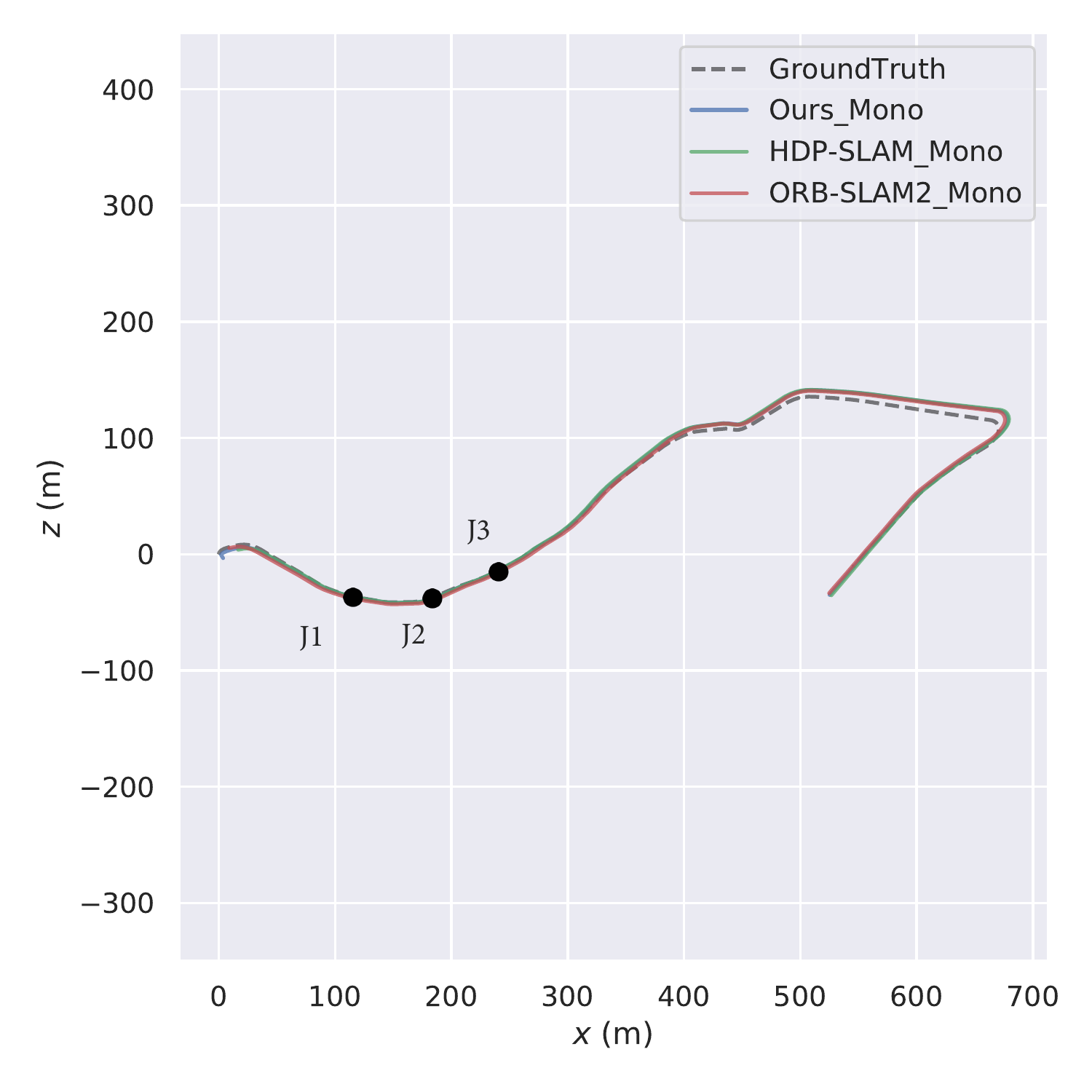}} \qquad
\centering 
\subfigure[Comparison of trajectories near point J1.]{ 
  
\includegraphics[width=1.3in, height=1.3in, trim=10 10 10 10, clip]{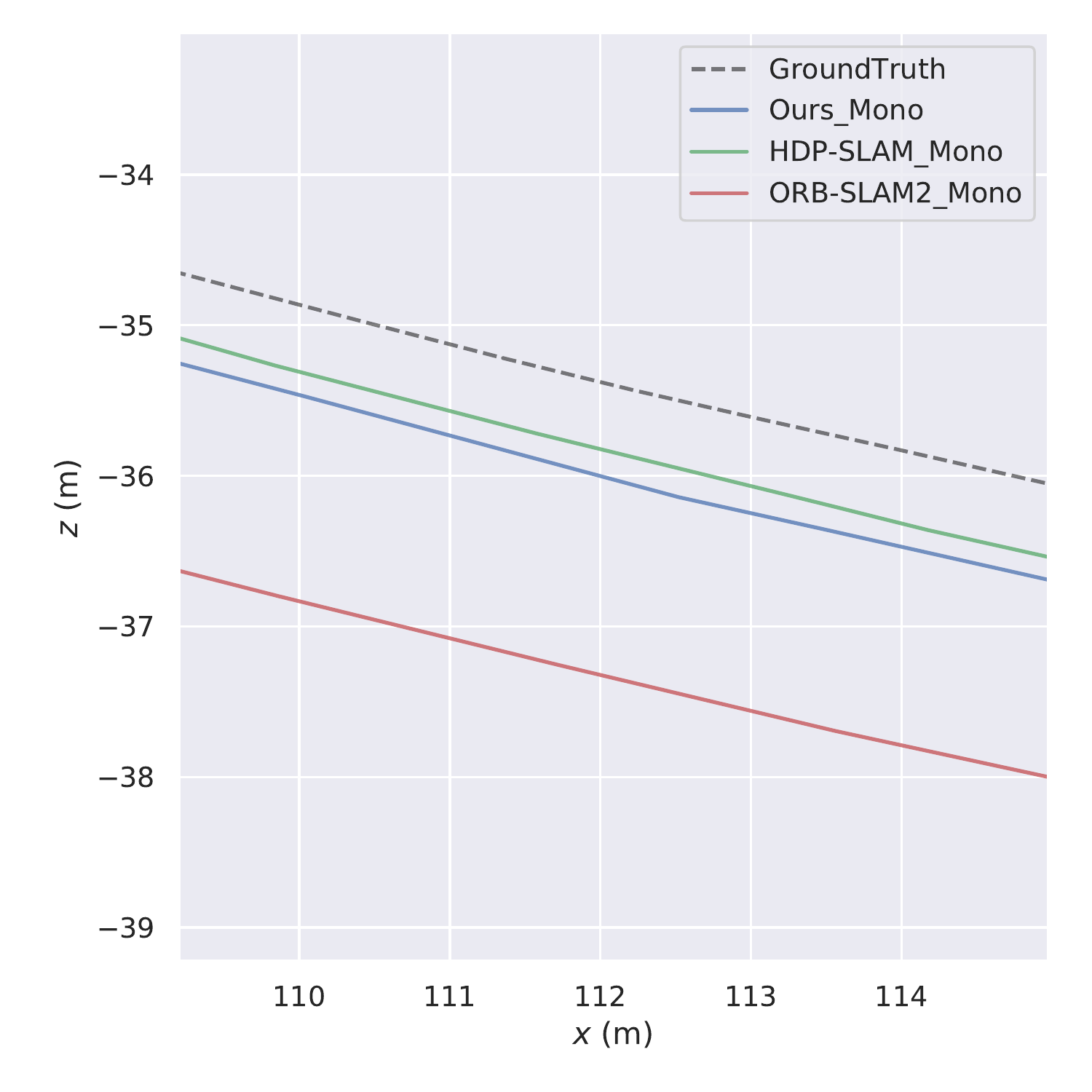}} \qquad
\centering 
\subfigure[Comparison of trajectories near point J2.]{ 
  
\includegraphics[width=1.3in, height=1.3in, trim=10 10 10 10, clip]{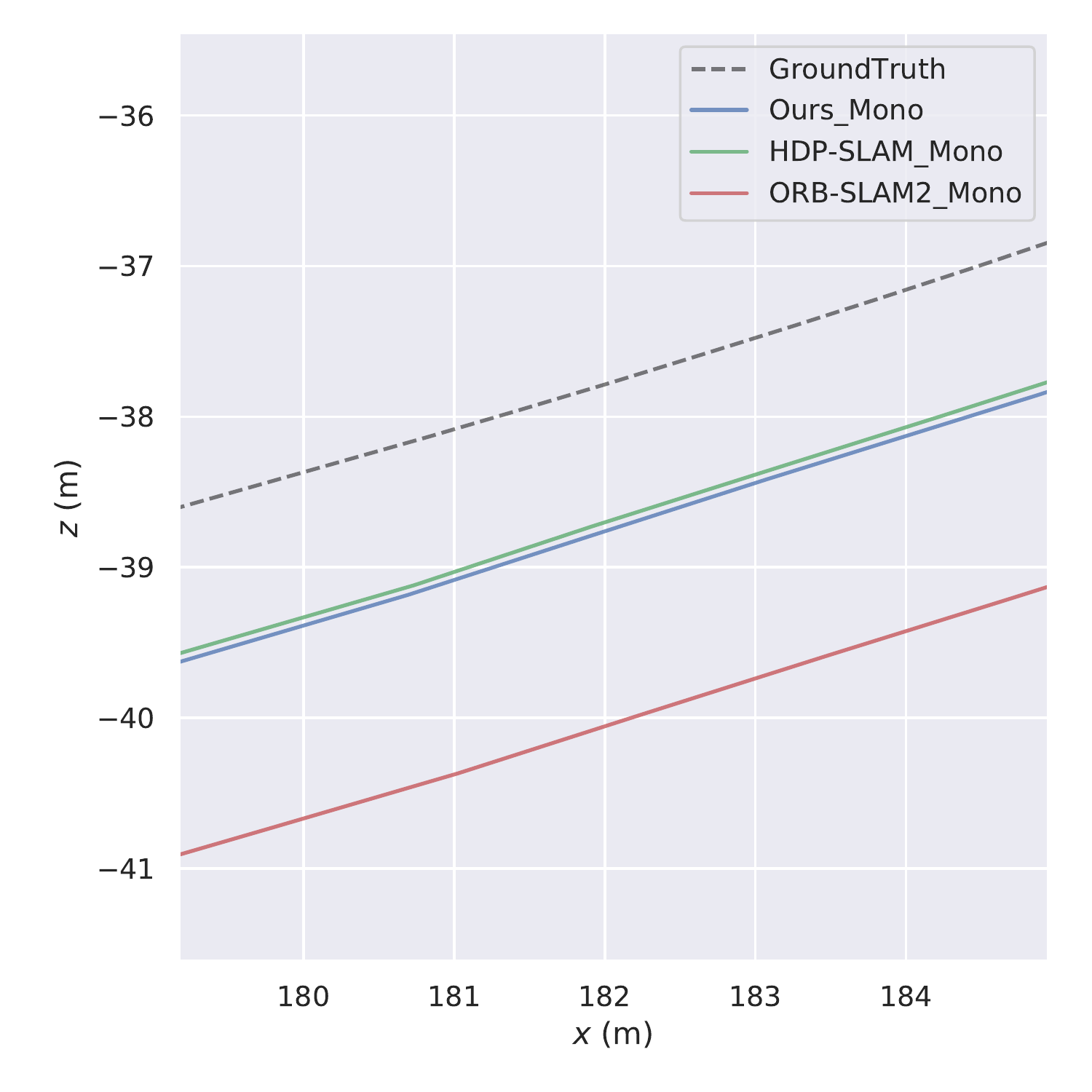}} \qquad
\centering  
\subfigure[Comparison of trajectories near point J3.]{ 
  
\includegraphics[width=1.3in, height=1.3in, trim=10 10 10 10, clip]{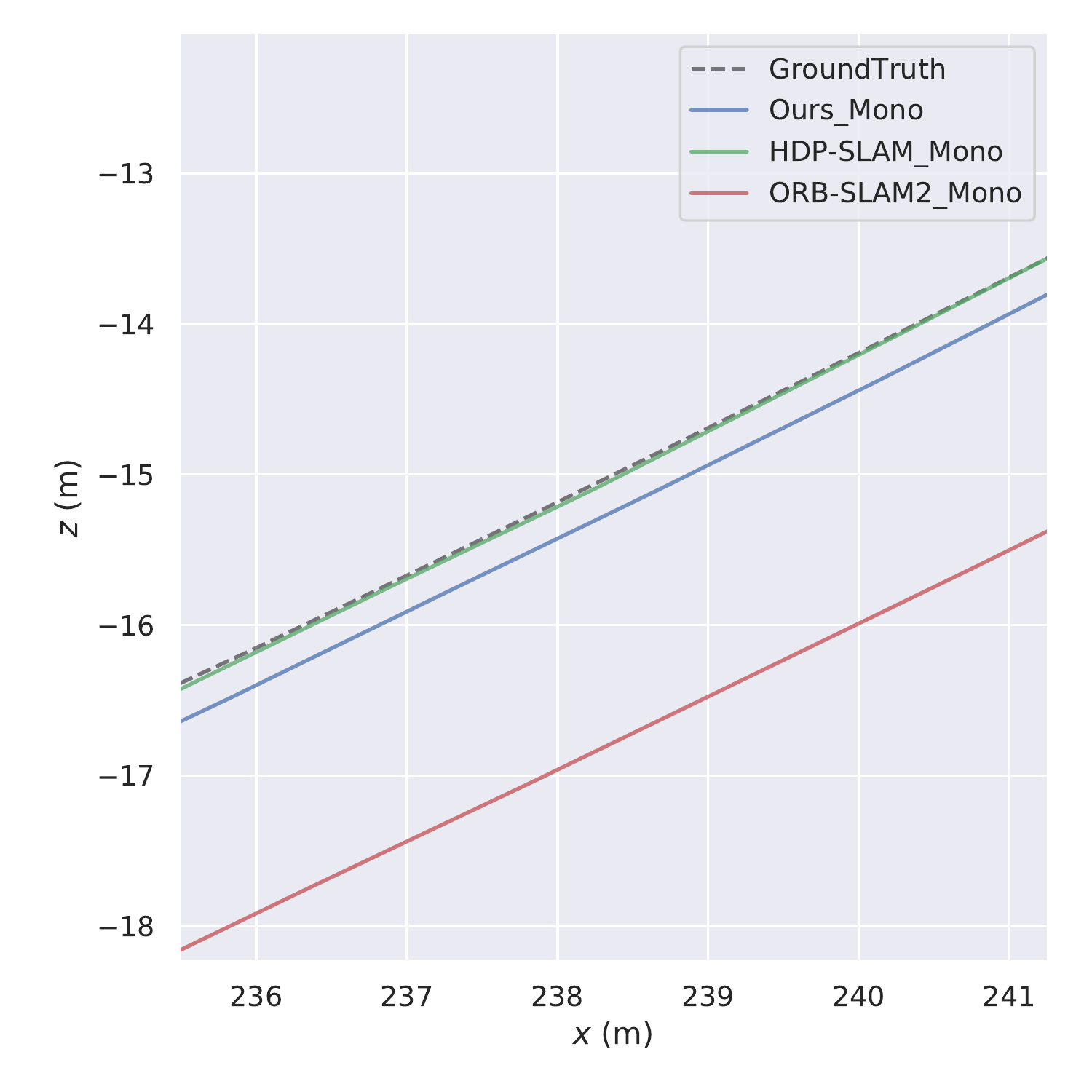}}

\caption{Comparison of trajectories in the KITTI dataset. Each row represents the comparison of one sequence. We show the results obtained in 03, 05, 07, and 10 sequences, respectively. Three points are randomly selected for the trajectory comparison of each sequence.} 
\label{Fig.6.} 
\end{figure*}

\subsection{Accuracy evaluation on the Simulation Hospital}\label{4.1}

In order to demonstrate the performance of our method on the simulation sequences, our method compares with the ORB-SLAM2 \cite{mur2017orb} and HDP-SLAM \cite{zhang2019hierarchical}.
These sequences are recorded in a simulated indoor hospital environment built by Microsoft AirSim\cite{shah2018airsim}, which is very close to the real-world environment. The sequence is composed of four scenes, namely the straight aisle, the left turn aisle, the right turn aisle, and a ward. The sequence aisle\_slow\_1 and aisle\_quick\_1 are recorded in the straight aisle scene, the sequence aisle\_slow\_2 and aisle\_quick\_2 are recorded in the left turn aisle scene, and the sequence aisle\_slow\_3 and aisle\_quick\_3 correspond to the right turn aisle scene, and the sequence office\_desk\_slow corresponds to the ward. The only difference between the corresponding aisle\_slow\_* and aisle\_quick\_* is the flying speed of the UAV. 
Therefore, when the flight path and shooting frequency are the same, the number of images captured at different flight speeds is different. The faster the flying speed of the UAV, the fewer the number of keyframe groups.

In order to effectively analyze our object association strategy, we compared our method from three aspects in Table \ref{tab1}. The first is the absolute trajectory error (ATE), which is one of the important indicators for SLAM performance evaluation. The second is the object association accuracy. The last is to compare the number of object landmarks in the map with the real number of objects in the sequence, which is represented by the column of Object\_GT. For all quantitative results, our method achieves the best one, due to more accurate object association and pose refinement.

Observing the results in the aisle\_slow\_1 and aisle\_quick\_1 sequences, we can find that the trajectory error result of the aisle\_slow\_1 is significantly better than that of the aisle\_quick\_1. We think the results are related to the flying speed of the UAV. When the flight path and camera shooting frequency are the same, the more images captured, the more semantic information obtained, and the pose optimization will be more accurate. However, as the number of object measurements increases, it becomes a challenge to object association so that the accuracy decreases. In the office\_desk\_slow sequence, we do not choose doors as landmarks, but select chairs as landmarks, which are similar in shape and size and are close to each other. Although we choose different landmarks, our approach still shows superiority. To make a fair comparison, we run each method ten times and take the average values as the final results.

Fig.\ref{Fig.1.} shows that the two doors are close together and have a similar appearance. Thus, they are very easy to be associated incorrectly in the process of object association. As shown in Fig.\ref{Fig.5.}, our method is less affected by this situation. This is partially attributed to the effect of a hierarchical grouping strategy and partially attributed to that the MOT can accurately associate object measurements in the short-term tracking. Since the trajectory errors of the three systems are at the centimeter level, we randomly select three points in Fig.\ref{Fig.5.}(c) and Fig.\ref{Fig.5.}(g)\footnote{The tool used to plot the result is from: github.com/MichaelGrupp/evo}, and zoom in the trajectory near these points to show the difference in detail. By comparing these trajectories, we can clearly see that trajectories of the proposed method are closer to the ground truth, which is achieved by the accurate object association and pose estimation.

According to these experimental results, the proposed hierarchical grouping and object association method is extremely useful and necessary. Besides, employing the MOT for object association within the keyframe group can essential improve the object detection rate.

\begin{table}[]
\caption{ Result of the object association between the proposed method and HDP-SLAM in the real hospital sequences. }
 \begin{center}
\setlength{\tabcolsep}{0.5mm}{
\begin{tabular}{|c|l|c|c|c|c|c|}
\hline
\multicolumn{2}{|c|}{\multirow{3}{*}{sequence number}} & \multicolumn{2}{c|}{Association accuracy(\%)} & \multicolumn{3}{c|}{Object number}                           \\ \cline{3-7} 
\multicolumn{2}{|c|}{}                                 & HDP-           & \multirow{2}{*}{Ours}         & \multirow{2}{*}{Object\_GT} & HDP-  & \multirow{2}{*}{Ours} \\
\multicolumn{2}{|c|}{}                                 & SLAM\cite{zhang2019hierarchical}          &                               &                             & SLAM\cite{zhang2019hierarchical} &                       \\ \hline
\multicolumn{2}{|c|}{hospital\_1}                      & 50.38         & \textbf{78.66}                & 17                          & 20   & \textbf{18}                    \\ \hline
\multicolumn{2}{|c|}{hospital\_2}                      & 53.39         & \textbf{80.80}                & 10                          & 12   & \textbf{10}                    \\ \hline
\multicolumn{2}{|c|}{hospital\_3}                      & 93.10         & \textbf{96.88}                & 6                           & 6    & \textbf{6}                     \\ \hline
\end{tabular}}
\end{center}
\label{tab2}  
\end{table}

\begin{table}[]
\caption{ Relative pose estimation error after returning to the start point. }
 \begin{center}
\setlength{\tabcolsep}{0.5mm}{
\begin{tabular}{|c|l|c|c|c|c|c|c|c|c|}
\hline
\multicolumn{2}{|c|}{\multirow{2}{*}{Method}} & \multirow{2}{*}{Roll} & \multirow{2}{*}{Pitch} & \multirow{2}{*}{Yaw} & Total          & \multirow{2}{*}{X} & \multirow{2}{*}{Y} & \multirow{2}{*}{Z} & Total          \\
\multicolumn{2}{|c|}{}                         &                       &                        &                      & Rot.(\degree)        &                    &                    &                    & Trans.(m)      \\ \hline
\multicolumn{2}{|c|}{ORB-SLAM2\cite{mur2017orb}}                & 0.786                 & 0.759                  & 0.799                & 1.354          & 2.324              & 0.263              & 1.691              & 2.886          \\ \hline
\multicolumn{2}{|c|}{HDP-SLAM\cite{zhang2019hierarchical}}                 & 0.008                 & \textbf{0.000}         & 0.010                & 0.013          & 0.388              & 0.134              & 0.389              & 0.565          \\ \hline
\multicolumn{2}{|c|}{Ours}                     & \textbf{0.006}        & 0.001                  & \textbf{0.009}       & \textbf{0.011} & \textbf{0.324}     & \textbf{0.127}     & \textbf{0.095}     & \textbf{0.361} \\ \hline
\end{tabular}}
\end{center}
\label{tab3}  
\end{table}

\begin{table*}[]
\caption{ Comparison of object association accuracy on KITTI sequence. }
\begin{center}
\setlength{\tabcolsep}{3mm}{\begin{tabular}{|c|l|c|l|c|c|c|l|c|}
\hline
\multicolumn{2}{|c|}{\multirow{3}{*}{sequence number}} & \multicolumn{3}{c|}{Association   accuracy(\%)}                              & \multicolumn{4}{c|}{Object size}                                                                     \\ \cline{3-9} 
\multicolumn{2}{|c|}{}                                 & \multicolumn{2}{c|}{\multirow{2}{*}{HDP-SLAM\cite{zhang2019hierarchical}}} & \multirow{2}{*}{Ours}       & \multirow{2}{*}{Object\_GT} & \multicolumn{2}{c|}{\multirow{2}{*}{HDP-SLAM\cite{zhang2019hierarchical}}} & \multirow{2}{*}{Ours} \\
\multicolumn{2}{|c|}{}                                 & \multicolumn{2}{c|}{}                          &                             &                             & \multicolumn{2}{c|}{}                          &                       \\ \hline
\multicolumn{2}{|c|}{KITTI-03}                         & \multicolumn{2}{c|}{48.53}                     & \textbf{63.43}              & 7                           & \multicolumn{2}{c|}{10}                        & \textbf{8}            \\ \hline
\multicolumn{2}{|c|}{KITTI-04}                         & \multicolumn{2}{c|}{62.00}                     & \textbf{73.81}              & 12                          & \multicolumn{2}{c|}{\textbf{11}}               & \textbf{11}           \\ \hline
\multicolumn{2}{|c|}{KITTI-05}                         & \multicolumn{2}{c|}{65.67}                     & \textbf{75.78}              & 92                          & \multicolumn{2}{c|}{94}                        & \textbf{89}           \\ \hline
\multicolumn{2}{|c|}{KITTI-06}                         & \multicolumn{2}{c|}{55.56}                     & \textbf{63.27}              & 66                          & \multicolumn{2}{c|}{42}                        & \textbf{64}           \\ \hline
\multicolumn{2}{|c|}{KITTI-07}                         & \multicolumn{2}{c|}{60.36}                     & \textbf{80.82}              & 105                         & \multicolumn{2}{c|}{78}                        & \textbf{92}           \\ \hline
\multicolumn{2}{|c|}{KITTI-10}                         & \multicolumn{2}{c|}{\textbf{77.03}}            & 73.80                       & 15                          & \multicolumn{2}{c|}{\textbf{22}}               & 24                    \\ \hline
\end{tabular}}
\end{center}
\label{tab6} 
\end{table*}

\begin{table*}[]
\caption{ Monocular Camera Pose Estimation Error on KITTI Odometry Benchmark.} 
\begin{center}
\setlength{\tabcolsep}{3mm}{
\begin{tabular}{|c|l|c|c|c|c|c|c|c|}
\hline
\multicolumn{9}{|c|}{Absolute Trajectory RMSE(m)}                                                                                              \\ \hline
\multicolumn{2}{|c|}{sequence number} & KITTI-03       & KITTI-04       & KITTI-05      & KITTI-06      & KITTI-07       & KITTI-10      & Mean           \\ \hline
\multicolumn{2}{|c|}{ORB-SLAM2\cite{mur2017orb}}       & 0.741          & 1.189          & 7.181         & 14.281        & 2.446          & 9.408         & 5.874          \\ \hline
\multicolumn{2}{|c|}{DynaSLAM\cite{bescos2018dynaslam}}        & 1.81           & 0.97           & \textbf{4.60} & 14.74         & 2.36           & \textbf{6.78} & 5.210          \\ \hline
\multicolumn{2}{|c|}{HDP-SLAM\cite{zhang2019hierarchical}}        & 0.595          & 0.931          & 6.983         & 12.594        & 1.878          & 8.957         & 5.323          \\ \hline
\multicolumn{2}{|c|}{CubeSLAM\cite{yang2019cubeslam}}        & 3.79           & 1.10           & 4.75          & \textbf{6.98} & 2.67           & 8.37          & 4.610          \\ \hline
\multicolumn{2}{|c|}{Dynamic-SLAM\cite{xiao2019dynamic}}    & 0.828          & 1.109          & 5.724         & 12.455        & 1.823          & 8.909         & 5.141          \\ \hline
\multicolumn{2}{|c|}{Ours}            & \textbf{0.312} & \textbf{0.673} & 5.329         & 10.503        & \textbf{1.815} & 9.005         & \textbf{4.606} \\ \hline
\end{tabular}}
\end{center}
\label{tab4} 
\end{table*}

\subsection{Accuracy evaluation on the real Hospital Dataset}\label{4.2}

We collect three sequences from a real hospital by hand-held ASUS Xtion2. The highest resolution can reach 640 $\times$ 480, so as to simulate the flight of UAVs in the inpatient department of this hospital. The length and width of the indoor environment are 50m and 30m, respectively. Two examples are shown in Fig.\ref{Fig.2.}. There are circular aisle, curved aisle, and straight aisle scenes, corresponding to hospital\_1, hospital\_2, and hospital\_3, respectively. 
In the sequence hospital\_1, the start point and the end point are at the same position. Thus, it can be used to quantitatively evaluate our method. We label a large number of door tags on these sequences.

Firstly, we verify our method from the aspect of object association. As shown in Table \ref{tab2}, the accuracy of the object association by the proposed method is more accurate than that of HDP-SLAM, and the number of objects is closer to the number of real objects. This means that the proposed method indeed improves the performance of semantic SLAM. Compared with the other two sequences, the hospital\_3 sequence has fewer people moving around and the light is brighter, so that the object association accuracy of the proposed method and the HDP-SLAM is above 90\%. 

Secondly, since the ground truth of trajectories cannot be obtained, we close the loop closing thread for all methods in the experiment on the hospital\_3 sequence, and compute the distance between the start and end points in the built map as the pose error. Experimental results are shown in Table \ref{tab3}, where we separately list rotational error (in degrees) and translational error (in meters). Compared with ORB-SLAM2, the HDP-SLAM and the proposed method have smaller pose errors between the start and end points. It is because two methods selectively and purposefully use semantic information to optimize the trajectory. More importantly, the proposed method uses a more stringent semantic filtering mechanism to achieve smaller pose errors between the starting point and the ending point.

Finally, according to these experimental results, our method is more robust than HDP-SLAM in a real hospital environment.

\subsection{System performance on KITTI Dataset}\label{4.3}

In order to better illustrate the generalization ability of the proposed method, we carry out experiments on 03, 04, 05, 06, 07 and 10 sequences in KITTI dataset. Similarly, we label lots of object tags on these sequences as ground truth. The above sequences in KITTI dataset contain many vehicles, and we set vehicles as landmarks on both sides of the road. By comparing the vehicle association and trajectory errors, the generalization performance and robustness of the system can be better evaluated.

In previous experiments, the RGB and depth images are as input to the SLAM system (RGBD-SLAM). Thus, the map points generated in the SLAM can be directed endowed with a centimeter-level depth. But in this part of experiments, we only use the monocular image in all SLAM systems, because there are only RGB image sequences in KITTI dataset without dense depth.

We first carry out experiments on the accuracy of object association. In Table \ref{tab6}, it is obvious that the object association accuracy of our semantic SLAM is much higher than that of HDP-SLAM. Only the association accuracy obtained from sequence 10 by the proposed method is slightly lower than that of HDP-SLAM. We think it is due to our method can detect more vehicles than HDP-SLAM. Pose optimization may be affected when a large number of object measurements are incorrectly associated.

Furthermore, we compare our method with state-of-the-art semantic SLAM systems in terms of trajectory errors. DynaSLAM\cite{bescos2018dynaslam}, HDP-SLAM\cite{zhang2019hierarchical}, CubeSLAM\cite{yang2019cubeslam} and Dynamic-SLAM\cite{xiao2019dynamic} also take semantic objects into account to reduce monocular scale drift. From the results in Table \ref{tab4}, our method achieves the best localization accuracy in most sequences. Through these sequences, we can prove that our method can be used in the monocular visual SLAM systems.

We further show the camera trajectories of the three methods in 03, 05, 07 and 10 sequences in Fig.\ref{Fig.6.}. We still randomly select three points in each of these sequences. By zooming in on the trajectories, we can see the superiority of the proposed method more vividly. In our method, we select the optimal target objects, which can effectively deal with the position and orientation of objects. This provides an important inspiration for us to join dynamic SLAM in our future work.

\section{Conclusion}
In this study, we propose a two-level hierarchical strategy for object association and an approach for optimizing the estimation of object poses, to enhance the accuracy and robustness of the semantic SLAM. The hierarchical object association integrates the short-term MOT-based local object association with the long-term HDP-based global object association. Thus, it can reduce association errors for object measurements corresponding to objects located closely and having similar appearances, and can achieve more robust semantic objects for the final semantic map. Moreover, the pose refinement approach provides more accurate object poses, which further improves the accuracy of trajectories.
Experimental results show that our method is more robust to ambiguous object association, and can achieve satisfactory performance in the hospital environment. In the future, we will consider elaborately dealing with the dynamic objects in scenes. And we also plan to incorporate the semantic SLAM with the path planning.


%


\ifCLASSOPTIONcaptionsoff
  \newpage
\fi



%


\bibliographystyle{IEEEtran}
\bibliography{ref}

\begin{thebibliography}{10}
\providecommand{\url}[1]{#1}
\csname url@samestyle\endcsname
\providecommand{\newblock}{\relax}
\providecommand{\bibinfo}[2]{#2}
\providecommand{\BIBentrySTDinterwordspacing}{\spaceskip=0pt\relax}
\providecommand{\BIBentryALTinterwordstretchfactor}{4}
\providecommand{\BIBentryALTinterwordspacing}{\spaceskip=\fontdimen2\font plus
\BIBentryALTinterwordstretchfactor\fontdimen3\font minus
  \fontdimen4\font\relax}
\providecommand{\BIBforeignlanguage}[2]{{%
\expandafter\ifx\csname l@#1\endcsname\relax
\typeout{** WARNING: IEEEtran.bst: No hyphenation pattern has been}%
\typeout{** loaded for the language `#1'. Using the pattern for}%
\typeout{** the default language instead.}%
\else
\language=\csname l@#1\endcsname
\fi
#2}}
\providecommand{\BIBdecl}{\relax}
\BIBdecl

\bibitem{yang2020combating}
G.~Yang, B.~Nelson, R.~Murphy, H.~Choset, H.~Christensen, S.~Collins, P.~Dario,
  K.~Goldberg, K.~Ikuta, N.~Jacobstein, D.~Kragic, R.~Taylor, and M.~McNutt,
  ``\BIBforeignlanguage{English (US)}{Combating covid-19-the role of robotics
  in managing public health and infectious diseases},''
  \emph{\BIBforeignlanguage{English (US)}{Science Robotics}}, vol.~5, no.~40,
  Mar. 2020.

\bibitem{bowman2017probabilistic}
S.~L. Bowman, N.~Atanasov, K.~Daniilidis, and G.~J. Pappas, ``Probabilistic
  data association for semantic slam,'' in \emph{2017 IEEE international
  conference on robotics and automation (ICRA)}.\hskip 1em plus 0.5em minus
  0.4em\relax IEEE, 2017, pp. 1722--1729.

\bibitem{mu2016slam}
B.~Mu, S.-Y. Liu, L.~Paull, J.~Leonard, and J.~P. How, ``Slam with objects
  using a nonparametric pose graph,'' in \emph{2016 IEEE/RSJ International
  Conference on Intelligent Robots and Systems (IROS)}.\hskip 1em plus 0.5em
  minus 0.4em\relax IEEE, 2016, pp. 4602--4609.

\bibitem{salas2013slam++}
R.~F. Salas-Moreno, R.~A. Newcombe, H.~Strasdat, P.~H. Kelly, and A.~J.
  Davison, ``Slam++: Simultaneous localisation and mapping at the level of
  objects,'' in \emph{Proceedings of the IEEE conference on computer vision and
  pattern recognition}, 2013, pp. 1352--1359.

\bibitem{rosinol2020kimera}
A.~Rosinol, M.~Abate, Y.~Chang, and L.~Carlone, ``Kimera: an open-source
  library for real-time metric-semantic localization and mapping,'' in
  \emph{2020 IEEE International Conference on Robotics and Automation
  (ICRA)}.\hskip 1em plus 0.5em minus 0.4em\relax IEEE, 2020, pp. 1689--1696.

\bibitem{choi2016local}
J.~Choi and M.~Maurer, ``Local volumetric hybrid-map-based simultaneous
  localization and mapping with moving object tracking,'' \emph{IEEE
  Transactions on Intelligent Transportation Systems}, vol.~17, no.~9, pp.
  2440--2455, 2016.

\bibitem{zhang2019hierarchical}
J.~Zhang, M.~Gui, Q.~Wang, R.~Liu, J.~Xu, and S.~Chen, ``Hierarchical topic
  model based object association for semantic slam,'' \emph{IEEE transactions
  on visualization and computer graphics}, vol.~25, no.~11, pp. 3052--3062,
  2019.

\bibitem{zhang2021fairmot}
Y.~Zhang, C.~Wang, X.~Wang, W.~Zeng, and W.~Liu, ``Fairmot: On the fairness of
  detection and re-identification in multiple object tracking,''
  \emph{International Journal of Computer Vision}, pp. 1--19, 2021.

\bibitem{shah2018airsim}
S.~Shah, D.~Dey, C.~Lovett, and A.~Kapoor, ``Airsim: High-fidelity visual and
  physical simulation for autonomous vehicles,'' in \emph{Field and service
  robotics}.\hskip 1em plus 0.5em minus 0.4em\relax Springer, 2018, pp.
  621--635.

\bibitem{geiger2013vision}
A.~Geiger, P.~Lenz, C.~Stiller, and R.~Urtasun, ``Vision meets robotics: The
  kitti dataset,'' \emph{The International Journal of Robotics Research},
  vol.~32, no.~11, pp. 1231--1237, 2013.

\bibitem{engel2014lsd}
J.~Engel, T.~Sch{\"o}ps, and D.~Cremers, ``Lsd-slam: Large-scale direct
  monocular slam,'' in \emph{European conference on computer vision}.\hskip 1em
  plus 0.5em minus 0.4em\relax Springer, 2014, pp. 834--849.

\bibitem{forster2014svo}
C.~Forster, M.~Pizzoli, and D.~Scaramuzza, ``Svo: Fast semi-direct monocular
  visual odometry,'' in \emph{2014 IEEE international conference on robotics
  and automation (ICRA)}.\hskip 1em plus 0.5em minus 0.4em\relax IEEE, 2014,
  pp. 15--22.

\bibitem{klein2007parallel}
G.~Klein and D.~Murray, ``Parallel tracking and mapping for small ar
  workspaces,'' in \emph{2007 6th IEEE and ACM international symposium on mixed
  and augmented reality}.\hskip 1em plus 0.5em minus 0.4em\relax IEEE, 2007,
  pp. 225--234.

\bibitem{mur2015orb}
R.~Mur-Artal, J.~M.~M. Montiel, and J.~D. Tardos, ``Orb-slam: a versatile and
  accurate monocular slam system,'' \emph{IEEE transactions on robotics},
  vol.~31, no.~5, pp. 1147--1163, 2015.

\bibitem{mur2017orb}
R.~Mur-Artal and J.~D. Tard{\'o}s, ``Orb-slam2: An open-source slam system for
  monocular, stereo, and rgb-d cameras,'' \emph{IEEE Transactions on Robotics},
  vol.~33, no.~5, pp. 1255--1262, 2017.

\bibitem{campos2021orb}
C.~Campos, R.~Elvira, J.~J.~G. Rodr{\'\i}guez, J.~M. Montiel, and J.~D.
  Tard{\'o}s, ``Orb-slam3: An accurate open-source library for visual,
  visual--inertial, and multimap slam,'' \emph{IEEE Transactions on Robotics},
  2021.

\bibitem{qiu2018endoscope}
L.~Qiu and H.~Ren, ``Endoscope navigation and 3d reconstruction of oral cavity
  by visual slam with mitigated data scarcity,'' in \emph{Proceedings of the
  IEEE Conference on Computer Vision and Pattern Recognition Workshops}, 2018,
  pp. 2197--2204.

\bibitem{marmol2019dense}
A.~Marmol, A.~Banach, and T.~Peynot, ``Dense-arthroslam: Dense intra-articular
  3-d reconstruction with robust localization prior for arthroscopy,''
  \emph{IEEE Robotics and Automation Letters}, vol.~4, no.~2, pp. 918--925,
  2019.

\bibitem{wang2007multi}
Z.~Wang, S.~Huang, and G.~Dissanayake, ``Multi-robot simultaneous localization
  and mapping using d-slam framework,'' in \emph{2007 3rd International
  Conference on Intelligent Sensors, Sensor Networks and Information}.\hskip
  1em plus 0.5em minus 0.4em\relax IEEE, 2007, pp. 317--322.

\bibitem{civera2011towards}
J.~Civera, D.~G{\'a}lvez-L{\'o}pez, L.~Riazuelo, J.~D. Tard{\'o}s, and J.~M.~M.
  Montiel, ``Towards semantic slam using a monocular camera,'' in \emph{2011
  IEEE/RSJ International Conference on Intelligent Robots and Systems}.\hskip
  1em plus 0.5em minus 0.4em\relax IEEE, 2011, pp. 1277--1284.

\bibitem{fang2021visual}
B.~Fang, G.~Mei, X.~Yuan, L.~Wang, Z.~Wang, and J.~Wang, ``Visual slam for
  robot navigation in healthcare facility,'' \emph{Pattern Recognition}, vol.
  113, p. 107822, 2021.

\bibitem{chen2020event}
G.~Chen, H.~Cao, J.~Conradt, H.~Tang, F.~Rohrbein, and A.~Knoll, ``Event-based
  neuromorphic vision for autonomous driving: a paradigm shift for bio-inspired
  visual sensing and perception,'' \emph{IEEE Signal Processing Magazine},
  vol.~37, no.~4, pp. 34--49, 2020.

\bibitem{wang2007d}
Z.~Wang, S.~Huang, and G.~Dissanayake, ``D-slam: A decoupled solution to
  simultaneous localization and mapping,'' \emph{The International Journal of
  Robotics Research}, vol.~26, no.~2, pp. 187--204, 2007.

\bibitem{lianos2018vso}
K.-N. Lianos, J.~L. Schonberger, M.~Pollefeys, and T.~Sattler, ``Vso: Visual
  semantic odometry,'' in \emph{Proceedings of the European conference on
  computer vision (ECCV)}, 2018, pp. 234--250.

\bibitem{mccormac2018fusion++}
J.~McCormac, R.~Clark, M.~Bloesch, A.~Davison, and S.~Leutenegger, ``Fusion++:
  Volumetric object-level slam,'' in \emph{2018 international conference on 3D
  vision (3DV)}.\hskip 1em plus 0.5em minus 0.4em\relax IEEE, 2018, pp. 32--41.

\bibitem{neira2001data}
J.~Neira and J.~D. Tard{\'o}s, ``Data association in stochastic mapping using
  the joint compatibility test,'' \emph{IEEE Transactions on robotics and
  automation}, vol.~17, no.~6, pp. 890--897, 2001.

\bibitem{strecke2019fusion}
M.~Strecke and J.~Stuckler, ``Em-fusion: Dynamic object-level slam with
  probabilistic data association,'' in \emph{Proceedings of the IEEE
  International Conference on Computer Vision}, 2019, pp. 5865--5874.

\bibitem{xiang2017rnn}
Y.~Xiang and D.~Fox, ``Da-rnn: Semantic mapping with data associated recurrent
  neural networks,'' \emph{arXiv preprint arXiv:1703.03098}, 2017.

\bibitem{liu2019monocular}
X.~Liu, S.~W. Chen, C.~Liu, S.~S. Shivakumar, J.~Das, C.~J. Taylor,
  J.~Underwood, and V.~Kumar, ``Monocular camera based fruit counting and
  mapping with semantic data association,'' \emph{IEEE Robotics and Automation
  Letters}, vol.~4, no.~3, pp. 2296--2303, 2019.

\bibitem{yang2019cubeslam}
S.~Yang and S.~Scherer, ``Cubeslam: Monocular 3-d object slam,'' \emph{IEEE
  Transactions on Robotics}, vol.~35, no.~4, pp. 925--938, 2019.

\bibitem{bavle2020vps}
H.~Bavle, P.~De~La~Puente, J.~P. How, and P.~Campoy, ``Vps-slam: visual planar
  semantic slam for aerial robotic systems,'' \emph{IEEE Access}, vol.~8, pp.
  60\,704--60\,718, 2020.

\bibitem{qin2020avp}
T.~Qin, T.~Chen, Y.~Chen, and Q.~Su, ``Avp-slam: Semantic visual mapping and
  localization for autonomous vehicles in the parking lot,'' in \emph{2020
  IEEE/RSJ International Conference on Intelligent Robots and Systems
  (IROS)}.\hskip 1em plus 0.5em minus 0.4em\relax IEEE, 2020, pp. 5939--5945.

\bibitem{li2020textslam}
B.~Li, D.~Zou, D.~Sartori, L.~Pei, and W.~Yu, ``Textslam: Visual slam with
  planar text features,'' in \emph{2020 IEEE International Conference on
  Robotics and Automation (ICRA)}.\hskip 1em plus 0.5em minus 0.4em\relax IEEE,
  2020, pp. 2102--2108.

\bibitem{qian2021semantic}
Z.~Qian, K.~Patath, J.~Fu, and J.~Xiao, ``Semantic slam with autonomous
  object-level data association,'' in \emph{2021 IEEE International Conference
  on Robotics and Automation (ICRA)}.\hskip 1em plus 0.5em minus 0.4em\relax
  IEEE, 2021, pp. 11\,203--11\,209.

\bibitem{chen2021pole}
G.~Chen, F.~Lu, Z.~Li, Y.~Liu, J.~Dong, J.~Zhao, J.~Yu, and A.~Knoll,
  ``Pole-curb fusion based robust and efficient autonomous vehicle localization
  system with branch-and-bound global optimization and local grid map method,''
  \emph{IEEE Transactions on Vehicular Technology}, 2021.

\bibitem{lee2016ground}
K.-H. Lee, J.-N. Hwang, G.~Okopal, and J.~Pitton,
  ``Ground-moving-platform-based human tracking using visual slam and
  constrained multiple kernels,'' \emph{IEEE transactions on intelligent
  transportation systems}, vol.~17, no.~12, pp. 3602--3612, 2016.

\bibitem{sanchez2016online}
R.~Sanchez-Matilla, F.~Poiesi, and A.~Cavallaro, ``Online multi-target tracking
  with strong and weak detections,'' in \emph{European Conference on Computer
  Vision}.\hskip 1em plus 0.5em minus 0.4em\relax Springer, 2016, pp. 84--99.

\bibitem{bewley2016simple}
A.~Bewley, Z.~Ge, L.~Ott, F.~Ramos, and B.~Upcroft, ``Simple online and
  realtime tracking,'' in \emph{2016 IEEE international conference on image
  processing (ICIP)}.\hskip 1em plus 0.5em minus 0.4em\relax IEEE, 2016, pp.
  3464--3468.

\bibitem{wojke2017simple}
N.~Wojke, A.~Bewley, and D.~Paulus, ``Simple online and realtime tracking with
  a deep association metric,'' in \emph{2017 IEEE international conference on
  image processing (ICIP)}.\hskip 1em plus 0.5em minus 0.4em\relax IEEE, 2017,
  pp. 3645--3649.

\bibitem{sun2019deep}
S.~Sun, N.~Akhtar, H.~Song, A.~Mian, and M.~Shah, ``Deep affinity network for
  multiple object tracking,'' \emph{IEEE transactions on pattern analysis and
  machine intelligence}, vol.~43, no.~1, pp. 104--119, 2019.

\bibitem{pang2020tubetk}
B.~Pang, Y.~Li, Y.~Zhang, M.~Li, and C.~Lu, ``Tubetk: Adopting tubes to track
  multi-object in a one-step training model,'' in \emph{Proceedings of the
  IEEE/CVF Conference on Computer Vision and Pattern Recognition}, 2020, pp.
  6308--6318.

\bibitem{liu2016ssd}
W.~Liu, D.~Anguelov, D.~Erhan, C.~Szegedy, S.~Reed, C.-Y. Fu, and A.~C. Berg,
  ``Ssd: Single shot multibox detector,'' in \emph{European conference on
  computer vision}.\hskip 1em plus 0.5em minus 0.4em\relax Springer, 2016, pp.
  21--37.

\bibitem{tulsiani2015viewpoints}
S.~Tulsiani and J.~Malik, ``Viewpoints and keypoints,'' in \emph{Proceedings of
  the IEEE Conference on Computer Vision and Pattern Recognition}, 2015, pp.
  1510--1519.

\bibitem{reynolds2009gaussian}
D.~A. Reynolds, ``Gaussian mixture models.'' \emph{Encyclopedia of biometrics},
  vol. 741, 2009.

\bibitem{bescos2018dynaslam}
B.~Bescos, J.~M. F{\'a}cil, J.~Civera, and J.~Neira, ``Dynaslam: Tracking,
  mapping, and inpainting in dynamic scenes,'' \emph{IEEE Robotics and
  Automation Letters}, vol.~3, no.~4, pp. 4076--4083, 2018.

\bibitem{xiao2019dynamic}
L.~Xiao, J.~Wang, X.~Qiu, Z.~Rong, and X.~Zou, ``Dynamic-slam: Semantic
  monocular visual localization and mapping based on deep learning in dynamic
  environment,'' \emph{Robotics and Autonomous Systems}, vol. 117, pp. 1--16,
  2019.

\end{thebibliography}


%


\begin{IEEEbiography}[{\includegraphics[width=1in, height=1.25in]{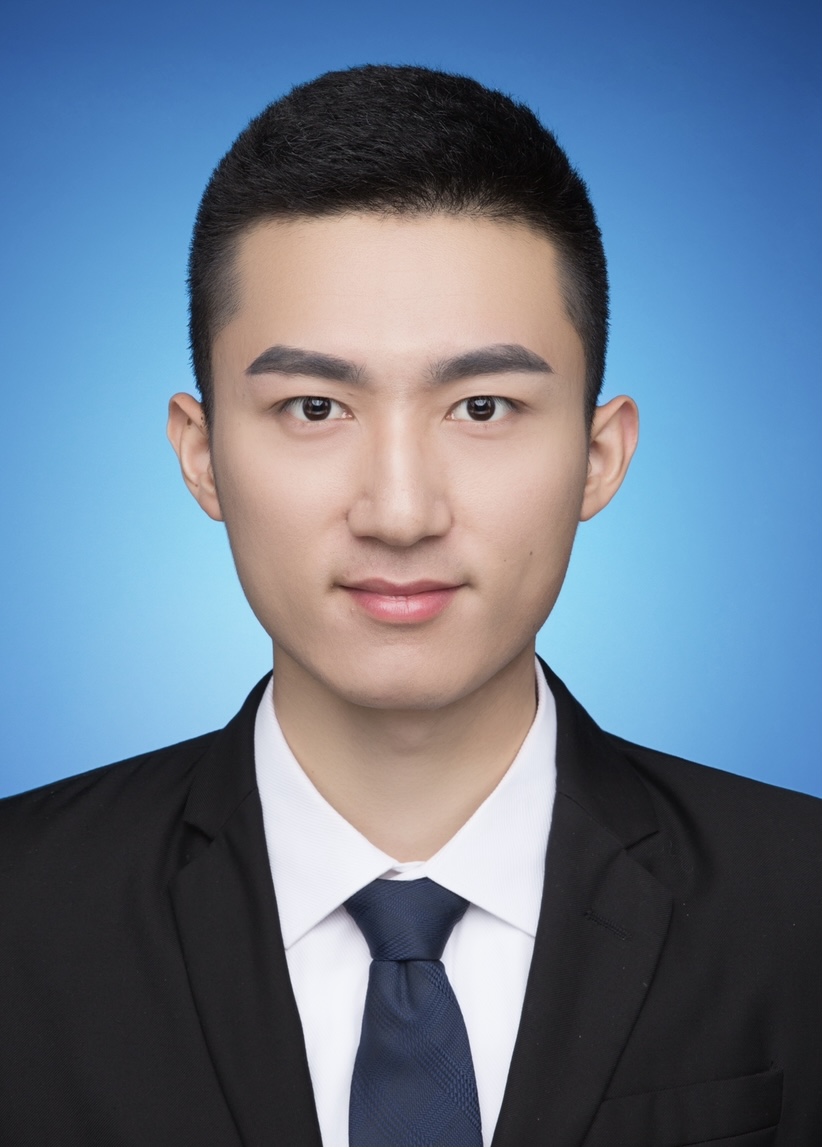}}]{Kaiqi Chen}
was born in Shaoxing, China. He received the B.S degree in software engineering from Southeast University Chengxian College, Nanjing, China, in 2019. He is currently a graduate student with Institute of Computer Vision, College of Computer Science and Technology, Zhejiang University of Technology. His current research focuses on object-based semantic SLAM and multi-object tracking.
\end{IEEEbiography}


\begin{IEEEbiography}[{\includegraphics[width=1in, height=1.25in] {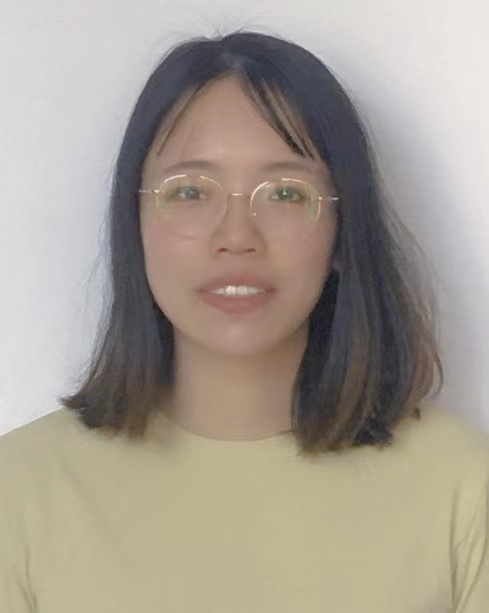}}]{Jialing Liu}
was born in Loudi, China. She received the B.S degree in software engineering from University of South China, Hengyang, China in 2019. She is currently a graduate student with Institute of Computer Vision, College of Computer Science and Technology, Zhejiang University of Technology. Her research interests include Multi-agent cooperation, and simultaneous localization and mapping (SLAM).
\end{IEEEbiography}

\begin{IEEEbiography}[{\includegraphics[width=1in, height=1.25in] {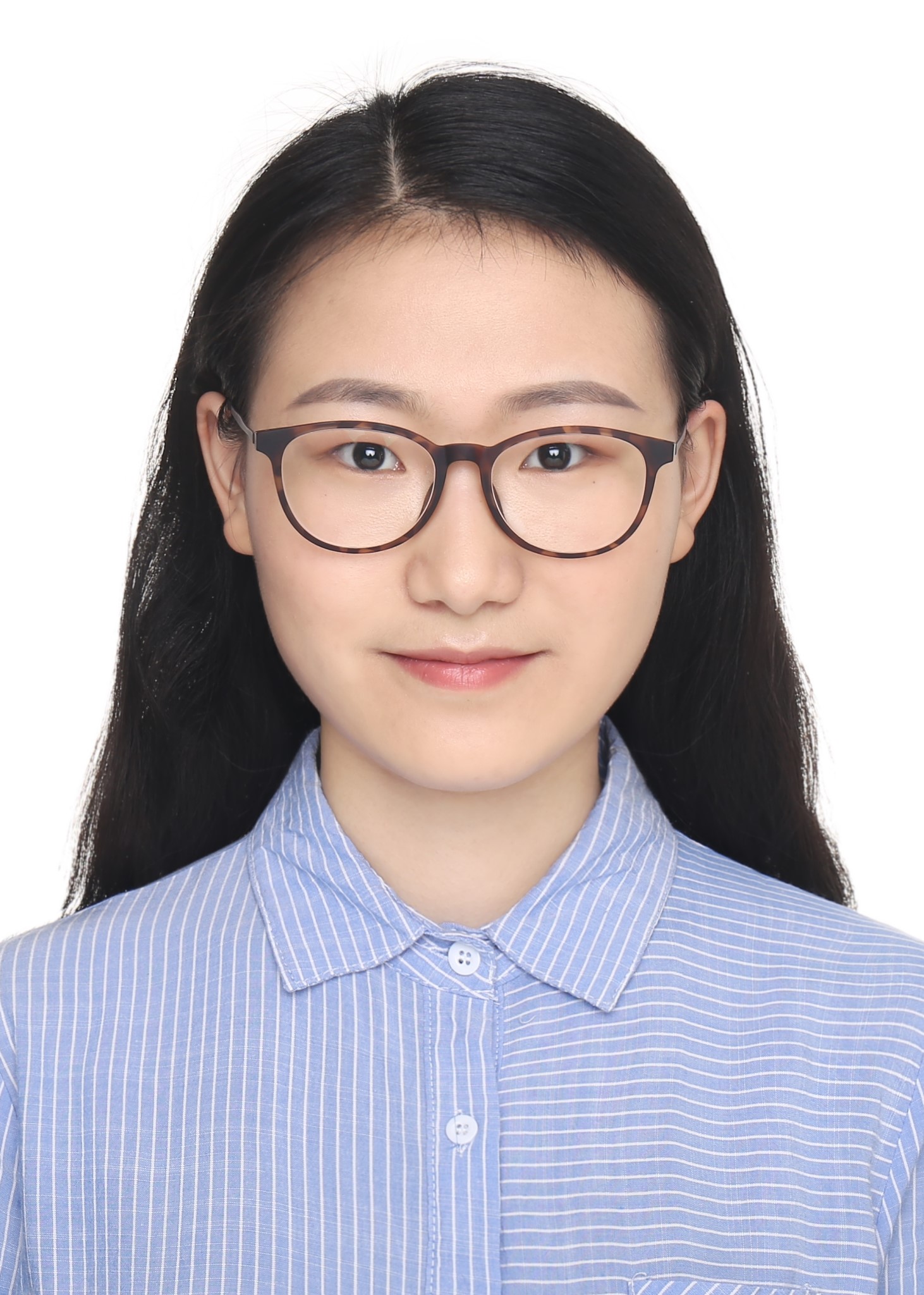}}]{Qinying Chen}
was born in Shaoxing, China. She received the B.S degree in computer science and technology from Zhejiang University of Technology, Hangzhou, China, in 2020.
She is currently a graduate student with Institute of Computer Vision, College of Computer Science and Technology, Zhejiang University of Technology. Her current research focuses on Multi-Sensor Fusion SLAM (simultaneous localization and mapping).
\end{IEEEbiography}

\begin{IEEEbiography}[{\includegraphics[width=1in, height=1.25in] {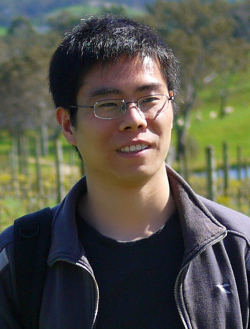}}]{Zhenhua Wang}
received the Ph.D. degree in computer vision from the University of Adelaide, Adelaide, SA, Australia, in 2014.
He is currently an assistant professor with the College of Computer Science and Technology, Zhejiang University of Technology, Hangzhou, China. His current research interests include computer vision, statistical learning, and pattern recognition.
\end{IEEEbiography}

\begin{IEEEbiography}[{\includegraphics[width=1in, height=1.25in] {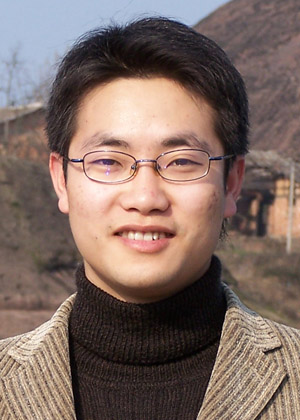}}]{Jianhua Zhang}
received the Ph.D. degree from University of Hamburg, Hamburg, Germany in 2012. He is currently a professor with the School of Computer Science and Engineering, Tianjin University of Technology, Tianjin, China. His current research interests include SLAM, 3D vision, reinforcement learning, and machine vision.
\end{IEEEbiography}








\end{document}